\documentclass[10pt,twocolumn,letterpaper]{article}

\usepackage{cvpr}              %

\definecolor{cvprblue}{rgb}{0.21,0.49,0.74}
\usepackage[pagebackref,breaklinks,colorlinks,allcolors=cvprblue]{hyperref}
\definecolor{mplblue}{rgb}{0.121569,0.466667,0.705882}
\def\paperID{} %
\def\confName{CVPR}
\def\confYear{2026}

\usepackage{hyperref}
\usepackage{url}
\usepackage{colortbl}
\usepackage{adjustbox}
\usepackage{subcaption}
\usepackage{multirow}
\usepackage{wrapfig}
\usepackage{dsfont} 
\usepackage{booktabs}
\usepackage{bm}
\usepackage{graphicx}
\usepackage{pgf}
\usepackage[colorinlistoftodos]{todonotes}
\usepackage{bbm}
\usepackage{import}
\usepackage{graphicx}
\usepackage{pgfplots}
\usepackage{xcolor} %
\usepackage{transparent} %
\usepackage{tikz}
\usetikzlibrary{arrows.meta, positioning, shapes.geometric}
\usepgfplotslibrary{groupplots}
\usetikzlibrary{calc}
\usepgfplotslibrary{fillbetween}
\pgfplotsset{compat=1.18}
\usepackage{lmodern}

\definecolor{mplblue}{rgb}{0.121569,0.466667,0.705882}
\usepackage[dvipsnames]{xcolor} 
\usepackage{amsmath,amssymb,amsfonts}
\usepackage{amsthm}
\usepackage{booktabs,tabularx}

\newcommand{\U}{\mathcal{U}}
\newcommand{\C}{\mathcal{C}}

\newcommand{\DL}{\mathcal{L}}

\newcommand{\ksig}{k_\sigma}

\newcommand{\appref}[1]{Appendix~\ref{#1}}
\newcommand{\tmpl}[1]{\texttt{``#1''}}

\title{Conformal Cross-Modal Active Learning}

\author{Huy Hoang Nguyen$^1$, Cedric Jung$^{1,2}$, Shirin Salehi$^3$, Tobias Glück$^1$, Anke Schmeink$^3$, Andreas Kugi$^{1,2}$\\
{\small $^1$ AIT Austrian Institute of Technology \quad
$^2$ Automation \& Control Institute, Technical University of Vienna} \\
{\small $^3$ Chair of Information Theory
and Data Analytics (INDA), RWTH Aachen University} \\
{\tt\small \{huy-hoang.nguyen,cedric.jung,tobias.glueck,andreas.kugi\}@ait.ac.at,} \\
{\tt\small \{shirin.salehi,anke.schmeink\}@inda.rwth-aachen.de}
}

\begin{document}
\maketitle

\begin{abstract}
Foundation models for vision have transformed visual recognition with powerful pretrained representations and strong zero-shot capabilities, yet their potential for data-efficient learning remains largely untapped. Active Learning (AL) aims to minimize annotation costs by strategically selecting the most informative samples for labeling, but existing methods largely overlook the rich multimodal knowledge embedded in modern vision–language models (VLMs).
We introduce \textbf{Conformal Cross-Modal Acquisition (CCMA)}, a novel AL framework that bridges vision and language modalities through a teacher–student architecture. CCMA employs a pretrained VLM as a teacher to provide semantically grounded uncertainty estimates, conformally calibrated to guide sample selection for a vision-only student model. By integrating multimodal conformal scoring with diversity-aware selection strategies, CCMA achieves superior data efficiency across multiple benchmarks. Our approach consistently outperforms state-of-the-art AL baselines, demonstrating clear advantages over methods relying solely on uncertainty or diversity metrics.

\end{abstract}

\section{Introduction}

In recent years, artificial intelligence has undergone a paradigm shift with the rise of foundation models, such as DALL-E \cite{ramesh_zero-shot_2021}, GPT-3 \citep{brown_language_2020}, Dinov2 \citep{oquab_dinov2_2024}, trained on broad data at scale. While these models provide powerful, transferable visual representations, their development requires massive amounts of curated data and computation resources. This challenge is especially pronounced in classification tasks, where large annotated datasets remain essential for high accuracy~\cite{evans2024bad}. To alleviate annotation and training costs, Active Learning (AL) has emerged as a compelling framework that aims to reduce annotation requirements by selecting the most informative samples for labeling~\citep{salehi2023active, salehi2023data, kaushal2019learning, li2024unlabeled}.

Pretrained visual features from foundation models have recently improved AL pipelines~\cite{gupte2024revisiting,chen2024making,tamkin2022active}, but most existing approaches remain \emph{vision-only}. Vision Language Models (VLMs) \citep{bordes_introduction_2024},
such as CLIP \citep{radford_learning_2021}, offer an untapped opportunity: their text–image alignment captures high-level class semantics, suggesting that they could provide more informative signals for sample selection than standard visual features alone. Preliminary attempts to use VLMs for AL focus mostly on prompt tuning \citep{bengar2021reducing, bang2024active}, leaving unexplored how to extract or quantify uncertainty from multimodal representations.

A key challenge arises: VLM outputs are often miscalibrated, domain-dependent, and not directly comparable to task-specific classifier probabilities, limiting their direct use as uncertainty oracles. Conformal Prediction (CP) \citep{angelopoulos2021gentle, shafer2008tutorial, silva2025conformal, nag2025conformal, conformalprediction} offers an appealing solution by providing distribution-free, per-sample uncertainty sets that remain valid regardless of model architecture or miscalibration. However, existing conformal AL methods~\cite{kharazian_copal_2024, sergio2020conformal} operate strictly within a single modality and do not leverage cross-modal semantic structure. Likewise, existing VLM-based AL does not use CP to fuse information from different models.

This gap motivates our research question: \textit{Can we incorporate the semantic structure of VLMs into an active learning acquisition
function using distribution-free conformal calibration?} We answer affirmatively by developing \textbf{Conformal Cross-Modal Acquisition (CCMA)}, a novel AL framework that bridges vision and language modalities through calibrated uncertainty estimation. CCMA employs a pretrained VLM as a teacher to generate semantically grounded prediction sets, which are conformally calibrated to provide distribution-free uncertainty estimates for guiding a vision-only student model. By integrating cross-modal conformal scoring with diversity-aware selection strategies, CCMA achieves superior data efficiency across multiple benchmarks.

Our main contributions are as follows:
\begin{enumerate}
    \item We propose a \textbf{teacher–student conformal scoring mechanism} that aligns vision-only predictions from a student model with text–image guidance from a pretrained VLM teacher. By constructing conformal prediction sets calibrated on held-out data, CCMA provides \emph{distribution-free, per-sample uncertainty} that is robust across datasets and architectures.
    
    \item We introduce a \textbf{selective subpooling strategy} based on clustering in CLIP~\cite{radford_learning_2021} feature space, which preserves geometric diversity while substantially reducing the number of candidates to be scored. Combined with an \textbf{uncertainty-weighted coverage} objective, CCMA achieves an effective trade-off between scalability and informativeness, enabling efficient active selection without accuracy degradation.
    
    \item We conduct extensive experiments on multiple image classification benchmarks, showing that CCMA consistently outperforms state-of-the-art active learning baselines across diverse domains and modalities.
    
    \item We provide a detailed analysis of the role of VLMs in active learning, revealing that CCMA excels when meaningful teacher–student discrepancies exist, while performance saturates once the teacher approaches oracle accuracy—transitioning the challenge from uncertainty estimation to coverage optimization.
\end{enumerate}
\begin{figure*}[t]
\centering
\def\svgwidth{\textwidth}
\import{figures/}{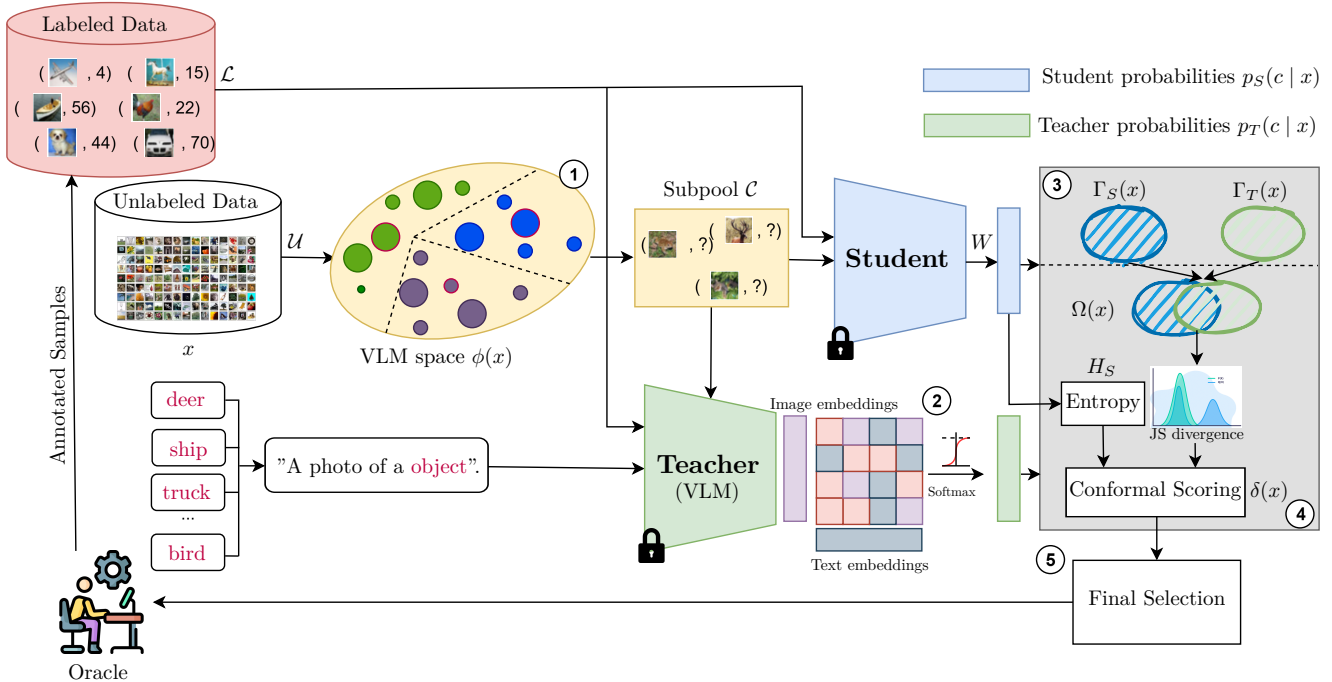}
\caption{\textbf{Overview of our proposed AL framework CCMA for image classification by exploring conformal prediction with multimodal uncertainty and diversity for efficient data acquisition}. Given labeled data $\mathcal{L}$ and an unlabeled pool $\mathcal{U}$, a frozen vision encoder (student) and a frozen VLM (teacher) serve as feature extractors, while a simple linear classifier is trained on the student features. \textbf{(1)} A selective subpool is formed via CLIP-space clustering. \textbf{(2–3)} Student and teacher posteriors $p_S$, $p_T$ are calibrated into conformal sets $\Gamma_S,\Gamma_T$. \textbf{(4)} Multimodal disagreement $\Omega(x)$ combines entropy and Jensen-Shannon (JS) divergence for uncertainty scoring $\delta(x)$. \textbf{(5)} Top-ranked samples are oversampled and clustered with uncertainty-weighted coverage for the final diverse selection.}
\label{fig:flowchart}
\end{figure*}
\section{Related Works}

\subsection{Active Learning}

In pool-based AL \citep{settles_active_2009,bald,houlsby_bayesian_2011} and classification tasks, the AL problem can be defined as follows: The whole dataset at first presents a small labeled dataset part named $\DL = \{(x_j, y_j)\}_{j=1}^M$ and a larger unlabeled part named $\U = \{x_i\}_{i=1}^N$, where $M \ll N$, $y_i \in \{0, 1\}$ is the class label of $x_i$ for binary classification, or $y_i \in \{1, \ldots, C\}$ for multi-class classification with $C$ classes. The process involves selecting instances from the unlabeled dataset $\U$ in a greedy manner, guided by a set of informativeness metrics called \textit{acquisition functions}. In each iteration $t$, a batch $\mathcal{D}_t^*$ of size $B$ from $\U$ is selected based on the learned model $\mathcal{M}$ and an acquisition function $\mathcal{A}(x, \mathcal{M})$, and queries their labels from the oracle. Data samples can be selected according to their acquisition score by $\mathcal{D}_t^* = {\overset{B}{\operatorname{argmax}}}_{x \in \U} \mathcal{A}(x, \mathcal{M})$, where the superscript $B$ indicates selection of the top $B$ points. In general, to reduce computational costs, a subpool $\mathcal{D}_{\text{pool}}$ is drawn from the unlabeled dataset $\U$, on which the acquisition function will be computed: $\mathcal{D}_t^* = {\overset{B}{\operatorname{argmax}}}_{x \in \mathcal{D}_{\text{pool}}} \mathcal{A}(x, \mathcal{M})$.

\noindent\textbf{Acquisition functions.} Uncertainty-based methods query unlabeled instances where the model is most uncertain. \textit{Entropy}~\cite{entropy} measures total predictive uncertainty by selecting samples with the highest entropy, while \textit{Margins}~\cite{margins} examines the gap between the top two class probabilities, selecting those with the smallest margin. Another \textit{Uncertainty} baseline~\cite{uncertainty} selects samples with the lowest maximum predicted probability. Beyond these point-estimate approaches, Bayesian active learning by disagreement (\textit{BALD})~\cite{bald} targets examples that maximize mutual information between predictions and the parameter posterior. \textit{PowerBALD}~\cite{powerbald} extends this by accounting for correlations among queried samples, thereby reducing the redundancy inherent in top-$B$ selection strategies. Most recently, \textit{UHerding}~\cite{uherding} maximizes an uncertainty-weighted coverage objective by using calibrated uncertainties and a shrinking kernel to balance diversity at low budgets with pure uncertainty at high budgets.

The choice between uncertainty-based and representativity/diversity-based acquisition functions reflects an exploration–exploitation trade-off, motivating hybrid approaches that combine or alternate between both~\cite{jung2024active, wang2025uncertainty}. Mixup-based methods, such as \textit{Alfa-mix}~\cite{alfamix}, interpolate unlabeled samples to create synthetic queries, whereas our approach operates directly in feature space and uses probabilistic set constructions to guide informative selection. Representative-based methods aim to cover the data distribution. Classical approaches such as the $k$-center greedy method~\cite{kcentergreedy} and \textit{Coreset}~\cite{coreset} minimize distances to cluster centers. Similarly, \textit{Typiclust}~\cite{typiclust} queries typical points within clusters, while \textit{ProbCover}~\cite{probcover} avoids outliers to improve representativeness. Leveraging clustering within a semantically meaningful feature space that is obtained via self-supervised learning further promotes diverse sampling.

\subsection{Conformal prediction}

Conformal prediction (CP)~\cite{conformalprediction} provides distribution-free, finite-sample uncertainty quantification by producing \emph{prediction sets} that contain the true label with user-specified coverage under the exchangeability assumption~\cite{shafer2008tutorial}. It has been explored in AL to quantify model uncertainty for regression~\citep{sergio2020conformal} and classification~\citep{makili_active_2012} tasks.
One approach ranks samples by conformal uncertainty, selecting those with the smallest $p$-values or the largest nonconformity (CoPAL~\cite{kharazian_copal_2024} and related CPAL variants~\cite{sergio2020conformal}). Another approach aims to reduce annotation costs by querying candidate class sets that are guaranteed to include the true label, prioritizing examples with small yet reliable sets~\cite{gwon2025enhancingcostefficiencyactive}. However, these methods are typically \emph{unimodal}, assigning each example a scalar based on a single model’s conformity or set size, while addressing diversity only heuristically.

\subsection{Active Learning in VLMs} %

Foundation models \citep{bordes_introduction_2024}, including vision–language models (VLMs), are trained on large unlabeled or noisy data, learning representations that enable strong zero- and few-shot performance. These properties make them natural candidates for active learning (AL), which seeks to maximize labeling efficiency. Although the two paradigms can complement each other~\citep{bengar2021reducing}, their integration remains mainly underexplored~\citep{gupte2024revisiting}. Yet, applying conventional AL frameworks to pretrained VLMs can sometimes degrade performance~\citep{bang2024active, yin2025towards}, motivating the development of AL strategies tailored to VLMs. For instance, \cite{safaei2025active} combines calibrated entropy with self- and neighbor-aware uncertainty to produce more reliable selection scores, narrowing the zero-shot–supervised gap of VLMs. Unlike their work, which applies AL for prompt tuning, we leverage cross-modal knowledge from VLM teachers to guide sample selection for vision-only students.

\noindent\textbf{Concluding,} while AL seeks to minimize annotation cost by selectively querying the most informative samples, its effectiveness is often hindered by unreliable uncertainty estimates from purely vision-based models. 
Recent VLMs provide semantically rich, cross-modal representations that can guide AL toward more meaningful and transferable sample selection. 
In this paper, we address this limitation by introducing a conformal prediction framework that bridges the uncertainty gap between visual and textual modalities, thereby enhancing both the efficiency and robustness of AL.

\section{Our method: Conformal Cross-Model Acquisition (CCMA)}

We propose a multi-modal conformal acquisition function that integrates diversity sampling with uncertainty estimation, using a student-teacher disagreement score. The approach follows a five-stage process, as depicted in Fig.~\ref{fig:flowchart}.

\subsection{Diverse subpool selection}\label{sec:subpool}

We adopt a \emph{compute-aware} candidate selection strategy by compressing the unlabeled pool $\U$ into a smaller candidate set $\C$ of size $|\mathcal{C}| \ll |\U|$. To construct $\C$, we cluster the VLM image embeddings ${\phi(x): x \in \U}$ in $\phi$-space (introduced in Sec.~\ref{sec:preds}) and choose one or a few representative points from each cluster using k-Means. 

We then apply the subsequent steps of CCMA exclusively to $\mathcal{C}$. This approach reduces the per-round scoring cost from $O(|\mathcal{U}|B)$ to $O(|\mathcal{C}|B)$ with $|\mathcal{C}|\ll|\mathcal{U}|$, while preserving coverage of the pool. Empirically, this diversity-selected subpool consistently outperforms an equally sized random subpool, yielding higher accuracy at low budgets and maintaining robustness without incurring the high computation of full-pool selection.

\subsection{Two predictors: student and teacher}\label{sec:preds}

For a given sample image $x$, we use image embeddings from pretrained VLM $\phi(x) \in \mathbb{R}^d$ and text prototypes $\{t_c\}_{c=1}^C \subset \mathbb{R}^d$, 
all $\ell_2$-normalized. 
The VLM teacher produces logits $\ell_T(c \mid x) = \phi(x)^\top t_c / \tau$, where $\tau$ is the temperature parameter. The posterior is then obtained via the softmax function:
\begin{equation}
    p_T(c \mid x) = \frac{\exp(\ell_T(c \mid x))}{\sum_{c'=1}^C \exp(\ell_T(c' \mid x))}.
\end{equation}
The student classifier $f_S$ operates on a separate backbone feature $z(x)\in\mathbb{R}^{D}$ (e.g., extracted from DINOv2) and outputs class probabilities $p_S(c \mid x)$.
The student consists of a feature adapter $\psi_\theta(\cdot)$ and a linear classification head with weight matrix
$W\in\mathbb{R}^{C\times D}$ with bias $b\in\mathbb{R}^C$, $h_c(x) \;=\; W\,\psi_\theta\!\big(z(x)\big)+b$, yielding:
\begin{equation}
  p_S(c\mid x) \;=\; \frac{\exp\big(h_c(x)\big)}{\sum_{c'=1}^C \exp\big(h_{c'}(x)\big)}.
  \label{eq:student-softmax}
\end{equation}
We train the parameters $\theta,W,$ and $b$ using cross-entropy loss on the labeled set and
evaluate the best checkpoint to obtain $\{p_S(c\mid x)\}_{c=1}^C$ for all $x$ in the candidate subpool.

\subsection{Two calibrated set predictors (split conformal)}\label{sec:sets}

For conformal prediction calibration, we define the nonconformity score as 
$a_m(x,c) = -\log p_m(c \mid x),$
where $p_m(c \mid x)$ denotes the predicted probability of class $c$ given input $x$, and $m \in \{T, S\}$ refers to the teacher T and student S. In a calibration split $\mathcal{C}_{\mathrm{cal}} \subseteq \mathcal{D}$, we determine thresholds $q_m$ for either the target expected set size $s_m$ or the marginal coverage $1-\alpha_m$. For size-targeted calibration, we find $q_m$ by bisection such that
\begin{equation}\label{eq:target-size}
\frac{1}{|\mathcal{C}_{\mathrm{cal}}|}\sum_{(x,y) \in \mathcal{C}_{\mathrm{cal}}} \left|\{c : a_m(x,c) \leq q_m\}\right| \approx s_m.
\end{equation}
For coverage-targeted calibration, we set $q_m$ as the empirical $(1-\alpha_m)$-quantile of nonconformity scores $\{a_m(x,y) : (x,y) \in \mathcal{C}_{\mathrm{cal}}\}$, guaranteeing split-conformal marginal coverage $\geq 1-\alpha_m$, where $\alpha_m \in [0, 1]$ represents the tolerated error rate. The set-valued predictors are then
\begin{equation}\label{eq:sets}
\Gamma_m(x) = \{c \in [C] : a_m(x,c) \leq q_m\}.
\end{equation}
We never force-add labels to $\Gamma_m(x)$; if sets are too small/large, we resolve $q_m$ to meet~\eqref{eq:target-size}. Because the calibration is split-conformal and applied independently to both modalities, this procedure is distribution-free and does not assume that the VLM teacher is well-calibrated or domain-aligned. Unlike prior AL methods that rely on raw VLM logits or treat VLMs as oracle predictors~\cite{safaei2025active}, our calibration guarantees valid finite-sample coverage for both teacher and student, making the cross-modal guidance robust even under severe teacher miscalibration or distribution shift.
\subsection{Cross-modal disagreement scoring}\label{sec:delta}

Given the conformal label sets $\Gamma_S(x)$ and $\Gamma_T(x)$, their union support is defined as $\Omega(x) = \Gamma_S(x) \cup \Gamma_T(x) \subseteq \{1,\dots,C\}$. For any posterior $p_m(\cdot \mid x) \in \Delta^{C-1}$, with $m \in \{T, S\}$, the distribution is renormalized over $\Omega(x)$:
\begin{equation}
p^{\Omega}_m(c\mid x)
\;=\;
\frac{p_m(c\mid x)\,\mathbbm{1}\{c\in\Omega(x)\}}
     {\sum_{c'\in\Omega(x)} p_m(c'\mid x)},
\end{equation}
with indicator function $\mathbbm{1}$. Once the renormalized posteriors $p^{\Omega}_T$ and $p^{\Omega}_S$ are obtained, the Jensen–Shannon (JS) divergence between them is computed as:
\begin{equation}\label{eq:js}
\begin{split}
\mathrm{JS}(p^{\Omega}_T\|p^{\Omega}_S)
&= \tfrac{1}{2}\,\mathrm{KL}\!\left(p^{\Omega}_T \,\middle\|\, 
\tfrac{p^{\Omega}_T+p^{\Omega}_S}{2}\right) \\
&\quad+\; \tfrac{1}{2}\,\mathrm{KL}\!\left(p^{\Omega}_S \,\middle\|\, 
\tfrac{p^{\Omega}_T+p^{\Omega}_S}{2}\right),
\end{split}
\end{equation}
where $\mathrm{KL}(p \,\|\, r)$ is the Kullback--Leibler divergence between distributions $p$ and $r$.
To dynamically balance the contributions of the student and teacher predictions, we introduce a parameter-free confidence gate. First, we define the top-1 confidence scores for the student and teacher models as
$\mathrm{conf}_S(x) = \max_{c} p_S(c \mid x),$ and $\mathrm{conf}_T(x) = \max_{c} p_T(c \mid x)$, respectively. Using these confidences, the confidence gate weight is computed as
\begin{equation}\label{eq:gate}
w_{\mathrm{js}}(x)
\;=\;
\frac{\mathrm{conf}_T(x)}
     {\mathrm{conf}_T(x) + \mathrm{conf}_S(x) + \varepsilon}
\;\in\;[0,1],
\end{equation}
where $\varepsilon>0$ is added for numerical stability. Intuitively, $w_{\mathrm{js}}(x)$ increases when the teacher is more confident than the student, thereby giving the teacher a greater influence on the final prediction, and vice versa. In early AL rounds, the teacher typically exhibits higher confidence and thus contributes more strongly to the score, whereas in later rounds, the student naturally takes over as its predictions sharpen. This adaptivity arises without additional hyperparameters and avoids committing to either a fully teacher-driven or student-driven rule, leading to the final score:
\begin{equation}\label{eq:final-score}
\begin{split}
\delta(x) &= w_{\text{js}}(x)\;\mathrm{JS}\!\Bigl(p_S^{\Omega}\,\Big\|\,p_T^{\Omega}\Bigr) \\
&\quad+\; \bigl(1-w_{\text{js}}(x)\bigr)\;H_S\!\bigl(y\mid x\bigr),
\end{split}
\end{equation}
where $H_S\!\bigl(y\mid x\bigr)$ is the entropy of the student's predictions:
\begin{equation}\label{eq:student-entropy}
H_S\!\bigl(y\mid x\bigr) \;=\; -\sum_{c=1}^{C} p_S(c\mid x)\,\log p_S(c\mid x).
\end{equation}
\subsection{Uncertainty-weighted coverage selection}\label{sec:greedy}

To construct a query batch of size $B$, we first define a candidate set
\[
S_{\kappa B} = \left\{\, x_i \in \mathcal{U} \;\middle|\; r_i \le \kappa B \,\right\}, 
\quad r_i = \operatorname{rank}_\downarrow(\delta(x_i))
\]
where $\delta(x_i)$ denotes the disagreement-based uncertainty score and $\kappa\ge1$ the oversampling factor.
Hence, $S_{\kappa B}$ contains the $\kappa B$ most uncertain samples, from which the final batch $S\subseteq S_{\kappa B}$, $|S|=B$, is selected by maximizing the \emph{uncertainty-weighted coverage}:
\begin{equation}\label{eq:objective}
F(S)\;=\;\frac{1}{|\U|}\sum_{u\in\U}\delta(u)\,\max_{s\in S}\ksig\!\big(\phi(u),\phi(s)\big),
\end{equation}
where $\ksig$ is a similarity kernel on the embedding space $\phi(\cdot)$.  The factor $\kappa$ trades off efficiency and accuracy: a larger $\kappa$ improves diversity and coverage at the cost of longer query time.

\section{Experiments}

\subsection{Experimental Setups}
\textbf{Datasets.} We evaluate our CCMA's performance against a suite of state-of-the-art AL methods across several benchmark datasets: CIFAR100~\cite{cifar100}, Food101~\cite{food101}, and DomainNet-Real~\cite{domainnetreal}, Caltech101~\cite{caltech101}, Caltech256~\cite{caltech256} (see~\appref{app:dataset-detail}).

\noindent\textbf{Implementation details.}
We employ a frozen CLIP~\cite{radford_learning_2021} ViT-L/14~\cite{vitl14} model as the \textbf{teacher}, where the text encoder provides class-wise prototypes for each downstream category using standard prompts such as ``A photo of a [CLS].'' (see~\appref{app:text-feature-extraction}). For each sample, logits are computed via temperature-scaled cosine similarity between image and text embeddings as described in Section~\ref{sec:preds}, with a fixed temperature $\tau=0.01$ (CLIP default) and $\tau=0.03$ for ablations. The \textbf{student} is a frozen vision-only backbone from DINOv2~\cite{oquab_dinov2_2024} followed by a linear classification head.

For active learning, we follow the training protocol of~\cite{gupte2024revisiting}, running $t=20$ iterations with five seeds \{1, 10, 100, 1000, 10000\}, and report the mean accuracy averaged over seeds. 
Unless otherwise stated, we fix the oversampling factor $\kappa=20$, target set sizes $s_T=3$ and $s_S=5$, and investigate the sensitivity to $\kappa$ in Section~\ref{sec:hypers}. Each query round acquires $B$ samples, e.g., $B=100$ for CIFAR100. Linear heads are trained using AdamW with a learning rate of $10^{-2}$, weight decay of $10^{-2}$, and dropout rate $\rho=0.75$.

\noindent\textbf{Baselines.}
We benchmark results using our method with \textit{eleven} AL baselines: random sampling (Random), uncertainty-based (Entropy, Margins, Uncertainty, BALD, BADGE, PowerBALD denoted by pBALD, Alfa-mix), representation-based (Coreset, Typiclust, ProbCover).

\subsection{Experimental Results}

\begin{figure}[!ht]
  \centering
  \begin{subfigure}[t]{\linewidth}
    \centering
    \def\svgwidth{\linewidth}
    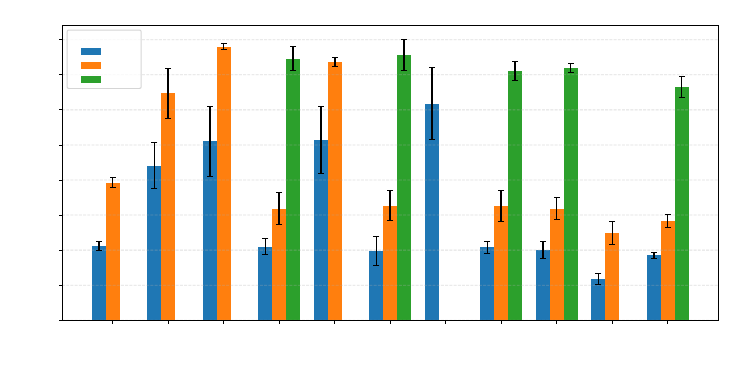
    \caption{Food101}
    \label{fig:labels-food101}
  \end{subfigure}
  \vspace{1em}
  \begin{subfigure}[t]{\linewidth}
    \centering
    \def\svgwidth{\linewidth}
    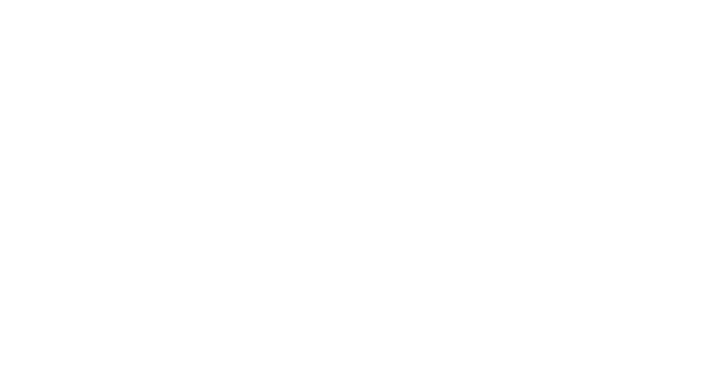
    \caption{DomainNet-Real}
    \label{fig:labels-domainnetreal}
  \end{subfigure}
  \caption{Labels required to reach target accuracies of 80\% (\textcolor{blue}{blue}), 85\% (\textcolor{orange}{orange}), and 90\% (\textcolor{ForestGreen}{green}) on Food101 and DomainNet-Real. Lower values indicate higher label efficiency. CCMA consistently reaches each accuracy threshold with fewer labeled samples than uncertainty- and coverage-based baselines, demonstrating improved sample efficiency across both datasets, especially in low-budget regimes.}
  \label{fig:labels-acc}
\end{figure}

\begin{table*}[!ht]
    \centering
    \caption{Mean accuracy averaged over 5 runs along with the standard deviation at AL iterations $t$ for datasets CIFAR100~\cite{cifar100}, Food101~\cite{food101}, and DomainNet-Real~\cite{domainnetreal} when utilizing the random initialization with DINOv2 ViT-g14 as the feature extractor. \textbf{Bold} values represent the \textbf{first-place} mean accuracy at iteration $t$, with the \underline{second-place} value \underline{underlined}.}
    \vspace{0.5em} 
    \begin{adjustbox}{width=1\textwidth}
    \begin{tabular}{c | c c c c c c c c c c c}
    \toprule 
        $t$ & Random & Uncertainty & Entropy & Margins & BALD & pBALD & Coreset & BADGE & Alfa-mix & ProbCover & \textit{CCMA (ours)} \\
    \midrule 
    \multicolumn{12}{c}{CIFAR100} \\
    \midrule
        1 & $48.0\pm 2.2$ & $48.0\pm 2.2$ & $48.0\pm 2.2$ & $48.0\pm 2.2$ & $48.0\pm 2.2$ & $48.0\pm 2.2$ & $48.0\pm 2.2$ & $48.0\pm 2.2$ & $48.0\pm 2.2$ & $48.0\pm 2.2$ & $48.0\pm 2.2$\\
        4 & $78.7\pm 1.6$ & $74.4\pm 1.8$ & $67.0\pm 1.2$ & $82.6\pm 1.1$ & $80.4\pm 0.5$ & $\underline{83.9}\pm 0.8$ & $77.9\pm 0.9$ & $\mathbf{84.1}\pm 0.6$ & $78.8\pm 1.4$ & $77.0\pm 5.8$ & $83.4\pm 1.2$\\
        8 & $85.8\pm 0.7$ & $84.2\pm 0.9$ & $80.7\pm 2.2$ & $87.9\pm 0.7$ & $85.2\pm 0.4$ & $88.4\pm 0.3$ & $84.3\pm 1.0$ & $\underline{88.5}\pm 0.4$ & $87.7\pm 0.8$ & $81.7\pm 4.6$ & $\mathbf{88.7}\pm 0.6$\\
        16 & $89.2\pm 0.2$ & $88.9\pm 0.8$ & $87.8\pm 0.5$ & $90.7\pm 0.2$ & $88.4\pm 0.3$ & $90.6\pm 0.2$ & $88.3\pm 0.8$ & $\underline{90.8}\pm 0.1$ & $90.6\pm 0.2$ & $86.3\pm 2.6$ & $\mathbf{91.1}\pm 0.4$\\
        20 & $89.8\pm0.2$ & $89.9\pm0.5$ & $89.4\pm0.1$ & $91.2\pm0.0$ & $89.6\pm0.3$ & $91.2\pm0.2$ & $88.9\pm0.6$ & $\underline{91.3}\pm0.1$ & $91.2\pm0.2$ & $88.0\pm1.8$ & $\mathbf{91.6}\pm0.2$\\
    \midrule
    \multicolumn{12}{c}{Food101} \\
    \midrule 
        1 & $46.8\pm1.9$ & $46.8\pm1.9$ & $46.8\pm1.9$ & $46.8\pm1.9$ & $46.8\pm1.9$ & $46.8\pm1.9$ & $46.8\pm1.9$ & $46.8\pm1.9$ & $46.8\pm1.9$ & $46.8\pm1.9$ & $46.8\pm1.9$ \\
        4 & $77.0\pm1.1$ & $63.0\pm3.2$ & $58.6\pm1.5$ & $75.4\pm2.1$ & $62.2\pm2.4$ & $\underline{78.1}\pm2.1$ & $62.8\pm3.7$ & $76.2\pm1.8$ & $77.7\pm1.6$ & $\mathbf{82.0}\pm0.9$ & $\underline{78.1}\pm0.9$\\
        8 & $83.8\pm0.3$ & $76.1\pm1.8$ & $73.0\pm2.4$ & $84.7\pm0.7$ & $73.8\pm2.3$ & $85.0\pm0.8$ & $72.8\pm2.9$ & $84.8\pm0.9$ & $85.1\pm1.1$ & $\underline{86.0}\pm0.5$ & $\mathbf{86.1}\pm0.5$\\
        16 & $87.4\pm0.4$ & $84.8\pm1.5$ & $81.9\pm1.5$ & $89.2\pm0.8$ & $82.3\pm1.7$ & $\underline{89.3}\pm0.3$ & $80.0\pm1.4$ & $\underline{89.3}\pm0.3$ & $89.1\pm0.2$ & $88.4\pm0.3$ & $\mathbf{90.1}\pm0.2$\\
        20 & $88.2\pm0.3$ & $86.4\pm1.0$ & $84.5\pm0.9$ & $90.0\pm0.1$ & $84.3\pm1.1$ & $90.1\pm0.3$ & $82.8\pm1.1$ & $\underline{90.2}\pm0.3$ & $\underline{90.2}\pm0.1$ & $88.7\pm0.5$ & $\mathbf{90.8}\pm0.2$\\
    \midrule
    \multicolumn{12}{c}{DomainNet-Real} \\
    \midrule 
        1 & $44.7\pm0.8$ & $44.7\pm0.8$ & $44.7\pm0.8$ & $44.7\pm0.8$ & $44.7\pm0.8$ & $44.7\pm0.8$ & $44.7\pm0.8$ & $44.7\pm0.8$ & $44.7\pm0.8$ & $44.7\pm0.8$ & $44.7\pm0.8$\\
        4 & $73.0\pm0.7$ & $64.0\pm1.1$ & $58.6\pm2.7$ & $72.9\pm0.5$ & $69.4\pm0.5$ & $75.7\pm0.6$ & $69.1\pm1.1$ & $75.5\pm0.6$ & $72.6\pm0.3$ & $\underline{76.3}\pm0.6$ & $\mathbf{77.7}\pm0.5$\\
        8 & $79.2\pm0.2$ & $74.3\pm0.6$ & $70.8\pm0.8$ & $79.4\pm0.5$ & $76.3\pm0.6$ & $\underline{80.8}\pm0.2$ & $75.7\pm0.4$ & $\underline{80.8}\pm0.1$ & $78.7\pm0.4$ & $78.9\pm0.8$ & $\mathbf{82.0}\pm0.2$\\
        16 & $82.1\pm0.3$ & $80.4\pm0.4$ & $79.0\pm0.4$ & $83.5\pm0.2$ & $80.9\pm0.4$ & $83.9\pm0.2$ & $80.3\pm0.5$ & $\underline{84.2}\pm0.2$ & $81.7\pm0.5$ & $79.5\pm0.5$ & $\mathbf{84.6}\pm0.2$\\
        20 & $82.8\pm0.1$ & $82.2\pm0.2$ & $80.9\pm0.3$ & $84.8\pm0.0$ & $82.1\pm0.3$ & $84.7\pm0.2$ & $81.4\pm0.6$ & $\underline{85.0}\pm0.1$ & $82.7\pm0.7$ & $79.7\pm0.5$ & $\mathbf{85.5}\pm0.1$\\
    \bottomrule
    \end{tabular}
    \end{adjustbox}
    \label{tab:ccma-accuracy}
\end{table*}

\noindent\textbf{Result analysis.}
Across all datasets, CCMA competitively outperforms existing active learning strategies in both early- and late-stage acquisition, demonstrating its effectiveness in leveraging cross-modal uncertainty for sample selection (Table~\ref{tab:ccma-accuracy}). On \textbf{CIFAR100}, CCMA achieves the highest final accuracy of $91.6\%$, surpassing the strongest baseline (BADGE) by $+0.3\%$ and maintaining a clear advantage throughout all acquisition rounds. \textbf{Food101} and \textbf{DomainNet-Real} show similar trends, with CCMA reaching $90.8\%$ and $85.5\%$, respectively, outperforming all competing uncertainty- and diversity-based methods, including BALD, pBALD, and Coreset. The gains are slightly improved in the early iterations ($t\le8$), where most baselines suffer from unstable uncertainty estimates, whereas CCMA benefits from the calibrated teacher–student disagreement, which provides more reliable per-sample confidence. Moreover, the improvements persist even in later rounds, indicating that CCMA does not merely focus on high-entropy regions but maintains semantic coverage through conformal calibration and diversity-aware selection. Notably, our conformal scoring adaptively balances the influence of teacher and student confidence, allowing the model to rely more on the student as its predictions become more reliable in later rounds, while still leveraging teacher guidance in early uncertain stages. 
This dynamic interplay enables CCMA to achieve a better balance between exploration (via subpool diversity) and exploitation (via multimodal uncertainty), leading to consistent gains across acquisition cycles.

\noindent\textbf{Label efficiency analysis.}
In Fig.~\ref{fig:labels-food101}, on \textbf{Food101}, CCMA and ProbCover are the only methods able to reach 85\% accuracy with fewer than 750 labeled samples. ProbCover performs competitively in the very low-budget regime due to its coverage-driven selection, which encourages early diversity and helps the model learn coarse class boundaries with minimal supervision. 
However, as the labeling budget increases, its lack of calibrated uncertainty limits further improvement. In contrast, CCMA maintains strong performance across all budget levels, the only method that achieves 90\% accuracy with fewer than 1.8K labeled samples, demonstrating its ability to exploit both uncertainty and diversity in a calibrated multimodal manner.

On the more challenging \textbf{DomainNet-Real} benchmark, which features significant domain variability and cross-category visual shifts, CCMA stands out as the only method capable of rapidly reaching 80\% accuracy with fewer than 2K labeled samples, as shown in Fig.~\ref{fig:labels-domainnetreal}. 
In contrast, other active learning baselines require substantially more annotations to achieve comparable performance. At higher accuracy thresholds, where many AL methods fail to progress beyond 85\%, CCMA, Badge, and PowerBALD remain the only approaches able to reach this level, albeit with larger budgets exceeding 6K samples. These results demonstrate that CCMA provides a favorable balance between early-stage label efficiency and sustained learning capacity, maintaining competitive performance even as the annotation budget increases. 

\subsection{When is the student better than the teacher in AL?}

\begin{figure}[h]
  \centering
  \includegraphics[width=0.95\linewidth]{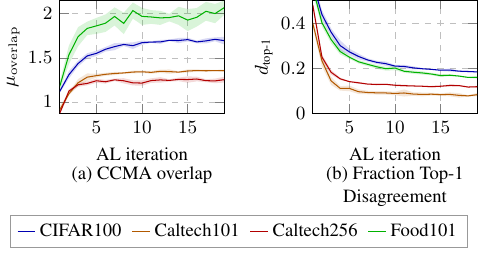}
  \caption{CCMA diagnostics in the overlap and the fraction Top-1 disagreement between teacher and student.}
  \label{fig:ccma-student}
\end{figure}

Modern active learning typically assumes a single learner, whereas CCMA introduces a teacher–student interaction in which a pretrained VLM teacher guides a vision-only student through conformal uncertainty calibration. But does this guidance always help, and under which conditions can the student surpass the teacher?

\begin{figure}[!ht]
  \centering
  \begin{subfigure}[t]{\linewidth}
    \includegraphics[width=0.97\linewidth]{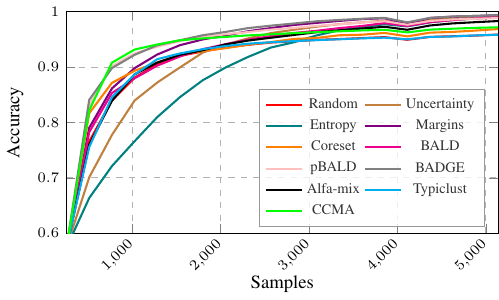}
    \caption{Caltech256}
    \label{fig:acc-caltech256}
  \end{subfigure}\vspace{0.3cm} 
  \begin{subfigure}[t]{\linewidth}
    \includegraphics[width=0.97\linewidth]{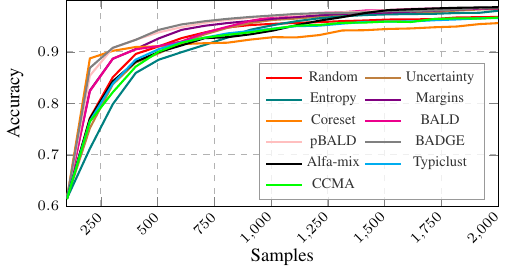}
    \caption{Caltech101}
    \label{fig:acc-caltech101}
  \end{subfigure}

  \caption{Test mean accuracy over 5 seeds for CCMA with other AL methods on Caltech256 and Caltech101 datasets.}
  \label{fig:caltech-acc}
\end{figure}

To address this, we analyze benchmarks using two diagnostics: mean CCMA overlap ($\mu_{\mathrm{overlap}} = \frac{1}{|\mathcal{U}|} \sum_{x \in \mathcal{U}} |\Gamma_T(x) \cap \Gamma_S(x)|$) and Top-1 disagreement fraction ($d_{\text{top-1}} = \frac{1}{|\mathcal{U}|} \sum_{x \in \mathcal{U}} 
\mathbbm{1}\!\left[\arg\max p_T(x) \neq \arg\max p_S(x)\right]$) shown in Fig.~\ref{fig:ccma-student}. On CIFAR100 and Food-101, the growing overlap and gradual decay of disagreement indicate that the teacher and student maintain complementary uncertainties across several rounds. CCMA capitalizes on this sustained mismatch, enabling more label-efficient learning and improved accuracy. In contrast, on Caltech101 and Caltech256, both overlap and disagreement flatten early, indicating that the student rapidly aligns with the teacher. Once this stage is reached, the teacher provides only a little new information, and the AL selection process becomes limited. As a result, uncertainty-based methods (e.g., BALD or BADGE) gain advantages in later rounds. This trend is reflected in the accuracy curves (Fig.~\ref{fig:caltech-acc}), where CCMA outperforms Random and other baselines, and achieves the highest accuracy on Caltech256 under low-budget settings ($<2$K samples) before plateauing as teacher–student disagreement collapses. In contrast, on Caltech101, CCMA remains competitive but only slightly exceeds Coreset and TypiClust, consistent with its coverage-limited nature.
Overall, these findings reveal that CCMA excels when meaningful teacher–student discrepancy persists—providing semantically grounded uncertainty signals that guide efficient exploration. Additional diagnostics, including JS divergence and confidence trends, are provided in the supplemental material for completeness (see~\appref{app:add-relation}).

\subsection{Impact of hyperparameter choice}\label{sec:hypers}

\begin{figure}[h]
  \centering
   \begin{subfigure}[t]{0.5\linewidth}
    \centering
  \includegraphics[width=0.97\linewidth]{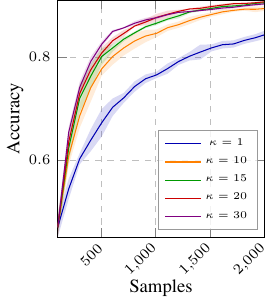}
    \caption{Food101}
    \label{fig:alpha-factor-food101}
  \end{subfigure}\hfill
  \begin{subfigure}[t]{0.5\linewidth}
    \includegraphics[width=0.97\linewidth]{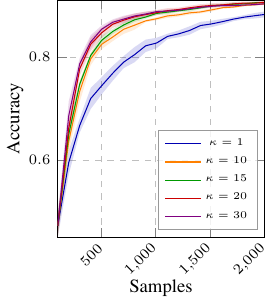}
    \caption{CIFAR100}
    \label{fig:alpha-factor-cifar100}
  \end{subfigure}
  \caption{Effect of oversampling factor $\kappa$.}
  \label{fig:alpha-factor}
\end{figure}

The experiments on Food101 and CIFAR100, shown in Fig.~\ref{fig:alpha-factor}, reveal that increasing $\kappa$ from 1 to 20 substantially enhances accuracy, especially when the sample size is small. Beyond $\kappa=20$, however, the performance gain plateaus, as higher values (e.g., 30) produce nearly identical accuracy curves. This suggests that $\kappa=20$ effectively captures the benefits of oversampling, with larger values offering negligible additional improvements.

\begin{figure}
  \centering
  \includegraphics[width=0.95\linewidth]{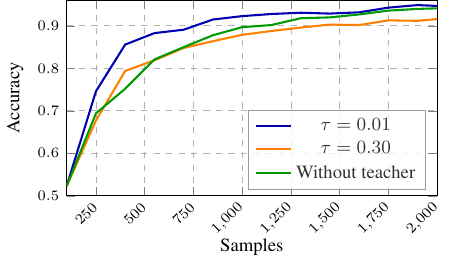}
  \caption{Motivation for investigating the impact of the teacher model in CCMA method. We report the CCMA's accuracy in three different regimes when the teacher is confident ($\tau=0.01$), weak ($\tau=0.30$), and disabled on the Food101 benchmark.}
  \label{fig:regimes-food101}
\end{figure}

The student’s performance is directly tied to the clarity of the teacher’s signals. Fig.~\ref{fig:regimes-food101} demonstrates that a lower temperature parameter ($\tau=0.01$) provides stronger, more informative supervision, enabling the student to achieve higher accuracy with fewer samples. Conversely, a higher $\tau$ ($\tau=0.30$) dilutes the teacher’s guidance.  However, as labeled size increases, the student trained without teacher guidance eventually reaches the same high-$\tau$ teacher’s performance. The student ultimately benefits from strong initial guidance, but may outgrow the teacher’s utility as the labeled set grows. This motivates CCMA’s hybrid rule, which uses teacher–student disagreement early and student entropy once the student becomes more reliable.

\subsection{Ablation study}
On CIFAR100, Food101, and DomainNet-Real, teacher–student mismatch is initially informative, and CCMA exploits it to achieve the best accuracy and label efficiency. Ablations show both subpool and final diversity matter, while a parameter-free confidence gate preserves performance without dataset-specific tuning. Query-time overhead is minimal, making CCMA a practical, data-aware active learner that knows when to trust the teacher and when to trust itself.

\begin{figure}
\centering
\includegraphics[width=\linewidth]{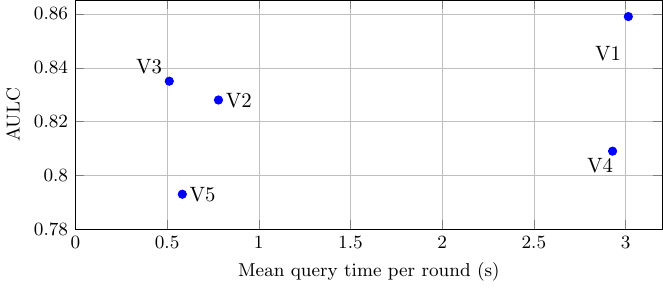}
\caption{AULC over round vs mean query time per round on the CIFAR100 dataset.}
\label{fig:aulc-timing}
\end{figure}

\noindent\textbf{Setup.} We evaluate five query variants on CIFAR100 (5 seeds; mean): 
V1 \textit{(ours)}—selective subpool + final diversity; 
V2—no subpool + final diversity; 
V3—random subpool + final diversity; 
V4—selective subpool, no diversity; 
V5—no subpool, no diversity. 
We report the metric Area Under Learning Curve (AULC)~\cite{reyes2018statistical} over rounds, along with the mean query time/round as shown in Fig.~\ref{fig:aulc-timing}.

Across all variants, the component that most directly lifts accuracy is the \emph{final diversity} stage. Removing only diversity (V4) depresses AULC from \textbf{0.859} to \textbf{0.809}, indicating that disagreement scoring alone is not sufficient to prevent redundancy in selected batches. By contrast, retaining diversity while altering the scope of scoring primarily affects \emph{efficiency}. Scoring the full pool with diversity (V2) is markedly faster than our default (V1), but the curated subpool in V1 yields a clear accuracy margin: +\textbf{0.031} AULC, at the cost of longer query time (3.02\,s). A large \emph{random} subpool with diversity (V3) provides a particularly attractive trade-off, achieving \textbf{0.835} AULC at only \textbf{0.51\,s}, which is roughly \textbf{6}\,$\times$ faster than V1 for a modest AULC loss of \textbf{2.4\%}. The ablations without diversity (V4, V5) consistently underperform their diversified counterparts (V2, V3), reinforcing that cross-modal disagreement must be paired with within-batch coverage to convert informative uncertainty into label efficiency. Overall, the evidence supports the view that \emph{diversity provides the accuracy gains} while \emph{curated subpooling controls scale}: our full method (V1) achieves the highest accuracy, while V3 provides the best option under strict time budgets.

\section{Conclusion}
We proposed \textbf{Conformal Cross-Modal Acquisition (CCMA)}, an active learning framework that combines conformal teacher--student uncertainty with diversity-aware selection for efficient dataset curation. 
By leveraging pretrained VLM guidance and calibrated prediction sets, CCMA provides more reliable selection signals than vision-only baselines, yielding strong label efficiency across diverse benchmarks. 
Our findings also reveal when multimodal guidance is beneficial versus when coverage dominates, offering actionable insight into deploying VLMs for real-world AL.

\noindent\textbf{Future work.}
We aim to extend CCMA to emerging-class and cross-domain settings, and to explore adaptive teacher--student co-learning for more scalable multimodal active learning.

{
    \small
    \bibliographystyle{ieeenat_fullname}
    \bibliography{main}

\begin{thebibliography}{49}
\providecommand{\natexlab}[1]{#1}
\providecommand{\url}[1]{\texttt{#1}}
\expandafter\ifx\csname urlstyle\endcsname\relax
  \providecommand{\doi}[1]{doi: #1}\else
  \providecommand{\doi}{doi: \begingroup \urlstyle{rm}\Url}\fi

\bibitem[Angelopoulos and Bates(2023)]{angelopoulos2021gentle}
Anastasios~N. Angelopoulos and Stephen Bates.
\newblock Conformal prediction: A gentle introduction.
\newblock \emph{Foundations and Trends® in Machine Learning}, 16\penalty0
  (4):\penalty0 494–591, 2023.

\bibitem[Ash and Adams(2020)]{ash2020warm}
Jordan Ash and Ryan~P Adams.
\newblock On warm-starting neural network training.
\newblock \emph{Advances in neural information processing systems (NeurIPS)},
  33:\penalty0 3884--3894, 2020.

\bibitem[Bae et~al.(2025)Bae, Oliveira, and Sutherland]{uherding}
Wonho Bae, Gabriel~L. Oliveira, and Danica~J. Sutherland.
\newblock Uncertainty herding: One active learning method for all label
  budgets.
\newblock In \emph{International Conference on Learning Representations
  (ICLR)}, 2025.

\bibitem[Bang et~al.(2024)Bang, Ahn, and Lee]{bang2024active}
Jihwan Bang, Sumyeong Ahn, and Jae-Gil Lee.
\newblock Active prompt learning in vision language models.
\newblock In \emph{IEEE/CVF Conference on Computer Vision and Pattern
  Recognition (CVPR)}, pages 27004--27014, 2024.

\bibitem[Bengar et~al.(2021)Bengar, van~de Weijer, Twardowski, and
  Raducanu]{bengar2021reducing}
Javad~Zolfaghari Bengar, Joost van~de Weijer, Bartlomiej Twardowski, and Bogdan
  Raducanu.
\newblock Reducing label effort: Self-supervised meets active learning.
\newblock In \emph{IEEE/CVF International Conference on Computer Vision
  (CVPR)}, pages 1631--1639, 2021.

\bibitem[Bordes et~al.(2024)Bordes, Pang, Ajay, Li, Bardes, Petryk, Ma{\~n}as,
  Lin, Mahmoud, Jayaraman, et~al.]{bordes_introduction_2024}
Florian Bordes, Richard~Yuanzhe Pang, Anurag Ajay, Alexander~C Li, Adrien
  Bardes, Suzanne Petryk, Oscar Ma{\~n}as, Zhiqiu Lin, Anas Mahmoud, Bargav
  Jayaraman, et~al.
\newblock An introduction to vision-language modeling.
\newblock \emph{arXiv preprint arXiv:2405.17247}, 2024.

\bibitem[Bossard et~al.(2014)Bossard, Guillaumin, and Van~Gool]{food101}
Lukas Bossard, Matthieu Guillaumin, and Luc Van~Gool.
\newblock Food-101 -- mining discriminative components with random forests.
\newblock In \emph{European Conference on Computer Vision (ECCV)}, pages
  446--461, 2014.

\bibitem[Brown et~al.(2020)Brown, Mann, Ryder, Subbiah, Kaplan, Dhariwal,
  Neelakantan, Shyam, Sastry, Askell, Agarwal, Herbert-Voss, Krueger, Henighan,
  Child, Ramesh, Ziegler, Wu, Winter, Hesse, Chen, Sigler, Litwin, Gray, Chess,
  Clark, Berner, McCandlish, Radford, Sutskever, and
  Amodei]{brown_language_2020}
Tom Brown, Benjamin Mann, Nick Ryder, Melanie Subbiah, Jared~D Kaplan, Prafulla
  Dhariwal, Arvind Neelakantan, Pranav Shyam, Girish Sastry, Amanda Askell,
  Sandhini Agarwal, Ariel Herbert-Voss, Gretchen Krueger, Tom Henighan, Rewon
  Child, Aditya Ramesh, Daniel Ziegler, Jeffrey Wu, Clemens Winter, Chris
  Hesse, Mark Chen, Eric Sigler, Mateusz Litwin, Scott Gray, Benjamin Chess,
  Jack Clark, Christopher Berner, Sam McCandlish, Alec Radford, Ilya Sutskever,
  and Dario Amodei.
\newblock Language models are few-shot learners.
\newblock In \emph{Advances in Neural Information Processing Systems
  (NeurIPS)}, pages 1877--1901, 2020.

\bibitem[Chen et~al.(2024)Chen, Bai, Huang, Lu, Wen, Yuille, and
  Zhou]{chen2024making}
Liangyu Chen, Yutong Bai, Siyu Huang, Yongyi Lu, Bihan Wen, Alan Yuille, and
  Zongwei Zhou.
\newblock Making your first choice: to address cold start problem in medical
  active learning.
\newblock In \emph{Medical Imaging with Deep Learning}, pages 496--525, 2024.

\bibitem[Dosovitskiy et~al.(2021)Dosovitskiy, Beyer, Kolesnikov, Weissenborn,
  Zhai, Unterthiner, Dehghani, Minderer, Heigold, Gelly, Uszkoreit, and
  Houlsby]{vitl14}
Alexey Dosovitskiy, Lucas Beyer, Alexander Kolesnikov, Dirk Weissenborn,
  Xiaohua Zhai, Thomas Unterthiner, Mostafa Dehghani, Matthias Minderer, Georg
  Heigold, Sylvain Gelly, Jakob Uszkoreit, and Neil Houlsby.
\newblock An image is worth 16x16 words: Transformers for image recognition at
  scale.
\newblock In \emph{International Conference on Learning Representations
  (ICLR)}, 2021.

\bibitem[Evans et~al.(2024)Evans, Pathak, Merzic, Schwarz, Tanno, and
  Henaff]{evans2024bad}
Talfan Evans, Shreya Pathak, Hamza Merzic, Jonathan Schwarz, Ryutaro Tanno, and
  Olivier~J Henaff.
\newblock Bad students make great teachers: Active learning accelerates
  large-scale visual understanding.
\newblock In \emph{European Conference on Computer Vision (ECCV)}, pages
  264--280, 2024.

\bibitem[Farahani and Hekmatfar(2009)]{kcentergreedy}
Reza~Zanjirani Farahani and Masoud Hekmatfar.
\newblock \emph{Facility location: Concepts, Models, algorithms and case
  studies}.
\newblock Springer Science \& Business Media, 2009.

\bibitem[Fei-Fei et~al.(2006)Fei-Fei, Fergus, and Perona]{caltech101}
Li Fei-Fei, Rob Fergus, and Pietro Perona.
\newblock One-shot learning of object categories.
\newblock \emph{IEEE Transactions on Pattern Analysis and Machine
  Intelligence}, 28\penalty0 (4):\penalty0 594--611, 2006.

\bibitem[Gal et~al.(2017)Gal, Islam, and Ghahramani]{bald}
Yarin Gal, Riashat Islam, and Zoubin Ghahramani.
\newblock Deep {B}ayesian active learning with image data.
\newblock In \emph{International Conference on Machine Learning (ICML)}, pages
  1183--1192, 2017.

\bibitem[Griffin et~al.(2007)Griffin, Holub, Perona, et~al.]{caltech256}
Gregory Griffin, Alex Holub, Pietro Perona, et~al.
\newblock Caltech-256 object category dataset.
\newblock Technical report, California Institute of Technology Pasadena, 2007.

\bibitem[Gupte et~al.(2024)Gupte, Aklilu, Nirschl, and
  Yeung-Levy]{gupte2024revisiting}
Sanket~Rajan Gupte, Josiah Aklilu, Jeffrey~J Nirschl, and Serena Yeung-Levy.
\newblock Revisiting active learning in the era of vision foundation models.
\newblock \emph{Transactions on Machine Learning Research (TMLR)}, 2024.

\bibitem[Gwon et~al.(2025)Gwon, Hwang, Kim, Ok, and
  Kwak]{gwon2025enhancingcostefficiencyactive}
Yeho Gwon, Sehyun Hwang, Hoyoung Kim, Jungseul Ok, and Suha Kwak.
\newblock Enhancing cost efficiency in active learning with candidate set
  query.
\newblock \emph{Transactions on Machine Learning Research (TMLR)}, 2025.

\bibitem[Hacohen et~al.(2022)Hacohen, Dekel, and Weinshall]{typiclust}
Guy Hacohen, Avihu Dekel, and Daphna Weinshall.
\newblock Active learning on a budget: Opposite strategies suit high and low
  budgets.
\newblock In \emph{International Conference on Machine Learning (ICML)}, pages
  8175--8195, 2022.

\bibitem[Houlsby et~al.(2011)Houlsby, Huszar, Ghahramani, and
  Lengyel]{houlsby_bayesian_2011}
Neil Houlsby, Ferenc Huszar, Zoubin Ghahramani, and Máté Lengyel.
\newblock Bayesian active learning for classification and preference learning.
\newblock \emph{Computing Research Repository (CoRR)}, 2011.

\bibitem[Jung et~al.(2024)Jung, Salehi, and Schmeink]{jung2024active}
C{\'e}dric Jung, Shirin Salehi, and Anke Schmeink.
\newblock Active learning with alternating acquisition functions: Balancing the
  exploration-exploitation dilemma.
\newblock In \emph{2024 IEEE International Conference on Big Data (BigData)},
  pages 5755--5764, 2024.

\bibitem[Kaushal et~al.(2019)Kaushal, Iyer, Kothawade, Mahadev, Doctor, and
  Ramakrishnan]{kaushal2019learning}
Vishal Kaushal, Rishabh Iyer, Suraj Kothawade, Rohan Mahadev, Khoshrav Doctor,
  and Ganesh Ramakrishnan.
\newblock Learning from less data: A unified data subset selection and active
  learning framework for computer vision.
\newblock In \emph{IEEE Winter Conference on Applications of Computer Vision
  (WACV)}, pages 1289--1299, 2019.

\bibitem[Kharazian et~al.(2024)Kharazian, Lindgren, Magnusson, and
  Bostr\"{o}m]{kharazian_copal_2024}
Zahra Kharazian, Tony Lindgren, Sindri Magnusson, and Henrik Bostr\"{o}m.
\newblock Copal: Conformal prediction in active learning an algorithm for
  enhancing remaining useful life estimation in predictive maintenance.
\newblock In \emph{Thirteenth Symposium on Conformal and Probabilistic
  Prediction with Applications}, pages 195--217, 2024.

\bibitem[Kirsch et~al.(2023)Kirsch, Farquhar, Atighehchian, Jesson,
  Branchaud-Charron, and Gal]{powerbald}
Andreas Kirsch, Sebastian Farquhar, Parmida Atighehchian, Andrew Jesson,
  Fr{\'e}d{\'e}ric Branchaud-Charron, and Yarin Gal.
\newblock Stochastic batch acquisition: A simple baseline for deep active
  learning.
\newblock \emph{Transactions on Machine Learning Research (TMLR)}, 2023.

\bibitem[Krizhevsky(2009)]{cifar100}
Alex Krizhevsky.
\newblock Learning multiple layers of features from tiny images.
\newblock 2009.

\bibitem[Lewis and Catlett(1994)]{uncertainty}
David~D. Lewis and Jason Catlett.
\newblock Heterogenous uncertainty sampling for supervised learning.
\newblock In \emph{International Conference on International Conference on
  Machine Learning (ICML)}, page 148–156, 1994.

\bibitem[Li et~al.(2024)Li, Wang, Chen, Lu, Fu, and Wu]{li2024unlabeled}
Xiongquan Li, Xukang Wang, Xuhesheng Chen, Yao Lu, Hongpeng Fu, and Ying~Cheng
  Wu.
\newblock Unlabeled data selection for active learning in image classification.
\newblock \emph{Scientific Reports}, 14\penalty0 (1):\penalty0 424, 2024.

\bibitem[Makili et~al.(2012)Makili, Sánchez, and
  Dormido-Canto]{makili_active_2012}
Lázaro~Emílio Makili, Jesús A.~Vega Sánchez, and Sebastián Dormido-Canto.
\newblock Active learning using conformal predictors: Application to image
  classification.
\newblock \emph{Fusion Science and Technology}, 62\penalty0 (2):\penalty0
  347--355, 2012.

\bibitem[Matiz and Barner(2020)]{sergio2020conformal}
Sergio Matiz and Kenneth~E. Barner.
\newblock Conformal prediction based active learning by linear regression
  optimization.
\newblock \emph{Neurocomputing}, 388:\penalty0 157--169, 2020.

\bibitem[Nag et~al.(2025)Nag, Ghosh, Ta, Bose, Li, and
  Roy-Chowdhury]{nag2025conformal}
Sayak Nag, Udita Ghosh, Calvin-Khang Ta, Sarosij Bose, Jiachen Li, and Amit~K
  Roy-Chowdhury.
\newblock Conformal prediction and mllm aided uncertainty quantification in
  scene graph generation.
\newblock In \emph{IEEE/CVF Conference on Computer Vision and Pattern
  Recognition (CVPR)}, pages 11676--11686, 2025.

\bibitem[Oquab et~al.(2024)Oquab, Darcet, Moutakanni, Vo, Szafraniec, Khalidov,
  Fernandez, HAZIZA, Massa, El-Nouby, Assran, Ballas, Galuba, Howes, Huang, Li,
  Misra, Rabbat, Sharma, Synnaeve, Xu, Jegou, Mairal, Labatut, Joulin, and
  Bojanowski]{oquab_dinov2_2024}
Maxime Oquab, Timoth{\'e}e Darcet, Th{\'e}o Moutakanni, Huy~V. Vo, Marc
  Szafraniec, Vasil Khalidov, Pierre Fernandez, Daniel HAZIZA, Francisco Massa,
  Alaaeldin El-Nouby, Mido Assran, Nicolas Ballas, Wojciech Galuba, Russell
  Howes, Po-Yao Huang, Shang-Wen Li, Ishan Misra, Michael Rabbat, Vasu Sharma,
  Gabriel Synnaeve, Hu Xu, Herve Jegou, Julien Mairal, Patrick Labatut, Armand
  Joulin, and Piotr Bojanowski.
\newblock {DINO}v2: Learning robust visual features without supervision.
\newblock \emph{Transactions on Machine Learning Research (TMLR)}, 2024.

\bibitem[Parvaneh et~al.(2022)Parvaneh, Abbasnejad, Teney, Haffari, van~den
  Hengel, and Shi]{alfamix}
Amin Parvaneh, Ehsan Abbasnejad, Damien Teney, Gholamreza~(Reza) Haffari, Anton
  van~den Hengel, and Javen~Qinfeng Shi.
\newblock Active learning by feature mixing.
\newblock In \emph{IEEE/CVF Conference on Computer Vision and Pattern
  Recognition (CVPR)}, pages 12237--12246, 2022.

\bibitem[Peng et~al.(2019)Peng, Bai, Xia, Huang, Saenko, and
  Wang]{domainnetreal}
Xingchao Peng, Qinxun Bai, Xide Xia, Zijun Huang, Kate Saenko, and Bo Wang.
\newblock Moment matching for multi-source domain adaptation.
\newblock In \emph{IEEE International Conference on Computer Vision (ICCV)},
  pages 1406--1415, 2019.

\bibitem[Radford et~al.(2021)Radford, Kim, Hallacy, Ramesh, Goh, Agarwal,
  Sastry, Askell, Mishkin, Clark, Krueger, and
  Sutskever]{radford_learning_2021}
Alec Radford, Jong~Wook Kim, Chris Hallacy, Aditya Ramesh, Gabriel Goh,
  Sandhini Agarwal, Girish Sastry, Amanda Askell, Pamela Mishkin, Jack Clark,
  Gretchen Krueger, and Ilya Sutskever.
\newblock Learning transferable visual models from natural language
  supervision.
\newblock In \emph{International Conference on Machine Learning (ICML)}, pages
  8748--8763, 2021.

\bibitem[Ramesh et~al.(2021)Ramesh, Pavlov, Goh, Gray, Voss, Radford, Chen, and
  Sutskever]{ramesh_zero-shot_2021}
Aditya Ramesh, Mikhail Pavlov, Gabriel Goh, Scott Gray, Chelsea Voss, Alec
  Radford, Mark Chen, and Ilya Sutskever.
\newblock Zero-shot text-to-image generation.
\newblock In \emph{International Conference on Machine Learning (ICML)}, pages
  8821--8831, 2021.

\bibitem[Reyes et~al.(2018)Reyes, Altalhi, and Ventura]{reyes2018statistical}
Oscar Reyes, Abdulrahman~H Altalhi, and Sebasti{\'a}n Ventura.
\newblock Statistical comparisons of active learning strategies over multiple
  datasets.
\newblock \emph{Knowledge-Based Systems}, 145:\penalty0 274--288, 2018.

\bibitem[Roy and McCallum(2001)]{margins}
Nicholas Roy and Andrew McCallum.
\newblock {Toward optimal active learning through sampling estimation of error
  reduction}.
\newblock In \emph{{International Conference on Machine Learning (ICML)}},
  pages 441--448, 2001.

\bibitem[Safaei and Patel(2025)]{safaei2025active}
Bardia Safaei and Vishal~M Patel.
\newblock Active learning for vision-language models.
\newblock In \emph{IEEE/CVF Winter Conference on Applications of Computer
  Vision (WACV)}, pages 4902--4912, 2025.

\bibitem[Salehi and Schmeink(2023{\natexlab{a}})]{salehi2023active}
Shirin Salehi and Anke Schmeink.
\newblock Is active learning green? an empirical study.
\newblock In \emph{IEEE International Conference on Big Data (BigData)}, pages
  3823--3829, 2023{\natexlab{a}}.

\bibitem[Salehi and Schmeink(2023{\natexlab{b}})]{salehi2023data}
Shirin Salehi and Anke Schmeink.
\newblock Data-centric green artificial intelligence: A survey.
\newblock \emph{IEEE Transactions on Artificial Intelligence}, 5:\penalty0
  1973--1989, 2023{\natexlab{b}}.

\bibitem[Sener and Savarese(2018)]{coreset}
Ozan Sener and Silvio Savarese.
\newblock Active learning for convolutional neural networks: A core-set
  approach.
\newblock In \emph{International Conference on Learning Representations
  (ICLR)}, 2018.

\bibitem[Settles(2009)]{settles_active_2009}
Burr Settles.
\newblock Active learning literature survey.
\newblock Computer Sciences Technical Report 1648, University of
  Wisconsin--Madison, 2009.

\bibitem[Shafer and Vovk(2008)]{shafer2008tutorial}
Glenn Shafer and Vladimir Vovk.
\newblock A tutorial on conformal prediction.
\newblock \emph{Journal of Machine Learning Research (JMLR)}, 9\penalty0 (3),
  2008.

\bibitem[Silva-Rodr{\'\i}guez et~al.(2025)Silva-Rodr{\'\i}guez, Ben~Ayed, and
  Dolz]{silva2025conformal}
Julio Silva-Rodr{\'\i}guez, Ismail Ben~Ayed, and Jose Dolz.
\newblock Conformal prediction for zero-shot models.
\newblock In \emph{IEEE/CVF Conference on Computer Vision and Pattern
  Recognition (CVPR)}, pages 19931--19941, 2025.

\bibitem[Tamkin et~al.(2022)Tamkin, Nguyen, Deshpande, Mu, and
  Goodman]{tamkin2022active}
Alex Tamkin, Dat Nguyen, Salil Deshpande, Jesse Mu, and Noah Goodman.
\newblock Active learning helps pretrained models learn the intended task.
\newblock \emph{Advances in Neural Information Processing Systems (NeurIPS)},
  35:\penalty0 28140--28153, 2022.

\bibitem[Vovk et~al.(2005)Vovk, Gammerman, and Shafer]{conformalprediction}
Vladimir Vovk, Alex Gammerman, and Glenn Shafer.
\newblock \emph{Algorithmic Learning in a Random World}.
\newblock Springer US, 2005.

\bibitem[Wang and Shang(2014)]{entropy}
Dan Wang and Yi Shang.
\newblock A new active labeling method for deep learning.
\newblock \emph{International Joint Conference on Neural Networks (IJCNN)},
  pages 112--119, 2014.

\bibitem[Wang and Zhao(2025)]{wang2025uncertainty}
Jiangyi Wang and Na Zhao.
\newblock Uncertainty meets diversity: A comprehensive active learning
  framework for indoor 3d object detection.
\newblock In \emph{IEEE/CVF Conference on Computer Vision and Pattern
  Recognition (CVPR)}, pages 20329--20339, 2025.

\bibitem[Yehuda et~al.(2022)Yehuda, Dekel, Hacohen, and Weinshall]{probcover}
Ofer Yehuda, Avihu Dekel, Guy Hacohen, and Daphna Weinshall.
\newblock Active learning through a covering lens.
\newblock \emph{Advances in Neural Information Processing Systems (NeurIPS)},
  35:\penalty0 22354--22367, 2022.

\bibitem[Yin et~al.(2025)Yin, Liu, and Sun]{yin2025towards}
Tianxiang Yin, Ningzhong Liu, and Han Sun.
\newblock Towards cost-effective learning: A synergy of semi-supervised and
  active learning.
\newblock In \emph{IEEE/CVF Conference on Computer Vision and Pattern
  Recognition (CVPR)}, pages 10163--10172, 2025.

\end{thebibliography}
}

\clearpage
\appendix

\twocolumn[{
\centering
\vspace*{0.5em}
{\LARGE \textbf{Conformal Cross-Modal Active Learning}\par}
\vspace{0.25em}
{\large Supplementary Material\par}
\vspace{1em}
}]

\section{Implementation details}
\label{app:dataset-detail}
 \textbf{CIFAR100} contains 60K natural images (32$\times$32) across 100 fine-grained categories, providing a standard benchmark for image classification under limited-label regimes.
\textbf{Food101} consists of 101K high-resolution food images with substantial intra-class variation.
\textbf{DomainNet-Real} is a subset of the large-scale DomainNet benchmark, with real-world images from 345 categories, and is used to assess robustness and cross-domain generalization.
\textbf{Caltech101} includes $\sim$9K images from 101 object categories with relatively clean backgrounds, while \textbf{Caltech256} extends this to 30K images over 256 categories with greater intra-class diversity.
At each active learning round, the acquisition batch size is set to the number of classes: $B=100$ for CIFAR100, $B=101$ for Food101, and $B=345$ for DomainNet-Real. We follow a fully \textit{cold-start} protocol~\cite{ash2020warm}: the student classifier is reinitialized and retrained from scratch at every acquisition step. Both the VLM teacher and the student’s vision backbone remain frozen throughout; features for all datasets are extracted once via a single forward pass and subsequently cached for all experiments.

\section{Text feature extraction from VLMs}
\label{app:text-feature-extraction}

For all experiments, the teacher model is a frozen CLIP ViT-L/14~\cite{radford_learning_2021} encoder that provides both image and text representations. 
To obtain class-wise text embeddings, we follow the CLIP text encoder pipeline and construct prompts for each dataset category using multiple natural language templates. 
Given a class name $c$, we build prompts of the form ``a photo of a \{class\}'', ``a bright photo of a \{class\}'', etc. 
Each prompt is tokenized and encoded with the CLIP text encoder, and the final class embedding is obtained by averaging the normalized embeddings across all templates.

\noindent\textbf{Prompt templates.}
Unless otherwise stated, we use the following default set:
\begin{tabularx}{\linewidth}{@{}X}
\tmpl{a photo of a \{class\}} \\
\tmpl{a photo of the \{class\}} \\
\tmpl{a blurry photo of a \{class\}}\\
\tmpl{a photo of one \{class\}} \\
\tmpl{a close-up photo of a \{class\}}\\
\tmpl{a rendition of a \{class\}} \\
\tmpl{a bright photo of a \{class\}} \\
\tmpl{a low resolution photo of a \{class\}} \\
\tmpl{a cropped photo of a \{class\}} \\
\tmpl{a clean photo of a \{class\}} \\
\end{tabularx}

These templates follow the standard CLIP zero-shot classification protocol~\cite{radford_learning_2021}, ensuring semantically rich and diverse text representations.

\vspace{0.25em}
\noindent\textbf{Dataset-specific class mappings.}
To handle datasets with different class label formats (e.g., numeric labels or textual names), we extract class names using a unified loader:
\begin{itemize}
    \item \textbf{CIFAR100, Caltech101, Caltech256, Food101:} class names are directly obtained from the torchvision library.
    \item \textbf{DomainNet-Real:} we normalize directory-based labels and map any non-alphanumeric identifiers via a lookup file.
    \item \textbf{Custom datasets:} if a mapping file (.txt) or explicit class list is provided, the loader automatically uses it; otherwise, class names are inferred from folder names or dataset metadata.
\end{itemize}

\vspace{0.25em}
\noindent\textbf{Feature storage.}
For each dataset, the resulting text features are saved under the same feature cache as image embeddings with shape $(C, d)$, where $C$ is the number of classes and $d$ is the CLIP embedding dimension. 
Each embedding is $\ell_2$-normalized and averaged across all templates, providing stable, semantically grounded class prototypes.

\section{Image feature extraction}
\label{app:image-feature-extraction}

To enable computationally efficient and reproducible active learning, CCMA operates entirely in a frozen feature space.
For all experiments, image embeddings are precomputed using pretrained vision backbones and stored in a shared cache. 
These embeddings are then used to train a simple linear classifier (student head) without updating the backbone weights.

\vspace{0.25em}
\noindent\textbf{Backbones.}
We employ pretrained models from standard vision foundations: DINOv2~\cite{oquab_dinov2_2024} for the student and CLIP ViT-L/14~\cite{radford_learning_2021} for the teacher.
The student model serves as the vision-only encoder producing embeddings $z(x)$, while the teacher’s CLIP image encoder produces multimodal-aligned embeddings $\phi(x)$.
Both backbones remain frozen throughout all active learning iterations.

\vspace{0.25em}
\noindent\textbf{Feature extraction and caching.}
Each dataset split (train/test) is passed through the frozen backbone once to extract image features.
Given an image $x_i$, the feature vector $f_i = \phi(x_i)$ is computed after the global pooling layer and $\ell_2$-normalized.
All extracted features and their class labels are stored in an HDF5 cache.
This allows CCMA to reuse embeddings across rounds, significantly reducing compute time.

\vspace{0.25em}
\noindent\textbf{Linear classifier training.}
During active learning, only a linear classification head $W \in \mathbb{R}^{C \times D}$ is trained on top of the frozen student features.
This design isolates the acquisition strategy from representation learning, ensuring that improvements stem purely from better sample selection rather than feature fine-tuning.

\section{Active learning training on the student model}
\label{app:student-training}

\noindent\textbf{Architecture.}
The student model is a simple linear classifier $h_c(x) \;=\; W\,\psi_\theta\!\big(z(x)\big)+b$ trained on top of the frozen DINOv2~\cite{oquab_dinov2_2024} features. 
The classifier includes an optional dropout layer for Monte Carlo sampling and uncertainty estimation. 
No gradients are propagated to the backbone, making each training round efficient and stable.

\vspace{0.25em}
\noindent\textbf{Training loop.}
During each active learning iteration:
\begin{enumerate}[leftmargin=1.5em, itemsep=1pt]
    \item The student model is trained on the currently labeled pool $\mathcal{L}$ using cross-entropy loss.
    \item Dropout-based sampling and conformal calibration yield per-sample predictive sets $\Gamma_S(x)$ and $\Gamma_T(x)$.
    \item The disagreement between teacher and student posteriors guides acquisition scoring.
    \item Selected samples are labeled and added to $\mathcal{L}$ for the next round.
\end{enumerate}

\vspace{0.25em}
\noindent\textbf{Prediction modes.}
The model supports multiple inference modes:
\begin{itemize}[leftmargin=1.2em,itemsep=1pt]
    \item \texttt{"probs"} – outputs softmax probabilities for uncertainty-based scoring.
    \item \texttt{"embed"} – returns intermediate embeddings for feature-space clustering.
    \item \texttt{"grad"} – computes gradient embeddings (e.g., for BADGE-style acquisition).
\end{itemize}
This flexibility allows fair comparisons with diverse active learning strategies while maintaining a consistent linear classifier backbone.

\section{Implementation details of split-conformal calibration}
\label{supp:conformal}

In the main paper (Section~\ref{sec:sets}), we introduced two calibrated set predictors (one for the teacher (\(m = T\)) and one for the student (\(m = S\))) based on split-conformal calibration.
Here, we provide additional details on the implementation, parameter settings, and rationale for the chosen target set sizes.

\noindent\textbf{Nonconformity scores and calibration.}
For both predictors, the nonconformity score is defined as
\[
a_m(x,c) = -\log p_m(c \mid x),
\]
where \(p_m(c \mid x)\) denotes the softmax probability assigned to class \(c\) by model $m \in \{T, S\}$.
A calibration split \(\mathcal{C}_{\mathrm{cal}}\) (disjoint from the active pool) is used to compute empirical thresholds \(q_m\) such that either the expected set size or the coverage constraint is satisfied.

\noindent\textbf{Size-targeted calibration.}
We target a desired mean prediction set size \(s_m\) by solving for the smallest threshold \(q_m\) that satisfies
\[
\frac{1}{|\mathcal{C}_{\mathrm{cal}}|} 
\sum_{(x,y) \in \mathcal{C}_{\mathrm{cal}}} 
\left|\{c : a_m(x,c) \leq q_m\}\right| 
\approx s_m.
\]
This threshold \(q_m\) is obtained via bisection search on the sorted nonconformity scores, ensuring the average conformal set size matches the target \(s_m\).

\noindent\textbf{Practical thresholds.}
In all experiments, we fix the target set sizes as:
\[
s_S = 5, \qquad s_T = 3.
\]
That is, the student predictor \(\Gamma_S(x)\) includes the top 5 most probable classes on average, while the teacher predictor \(\Gamma_T(x)\) includes the top 3.
These choices were determined empirically to balance informativeness and calibration stability:
- smaller \(s_T\) yields sharper and more confident teacher sets, focusing guidance on the most relevant semantic hypotheses;
- slightly larger \(s_S\) encourages exploration and helps the student capture broader class uncertainty during early rounds.

The calibrated thresholds \(q_m\) are recomputed at each active learning iteration using the current model’s predictions on the calibration split, ensuring adaptivity as the student improves.

\noindent\textbf{Coverage and reliability.}
Although our setup targets fixed expected set sizes, we verify that the resulting coverage (the fraction of samples whose true label lies in \(\Gamma_m(x)\)) remains high across rounds:
typically \(0.90\text{--}0.95\) for the teacher and \(0.85\text{--}0.90\) for the student.
This behavior is consistent with the conformal guarantee that for monotone \(a_m\), quantile-based thresholds yield calibrated coverage independent of data distribution.

In summary, our conformal calibration procedure introduces no additional learnable parameters and incurs negligible computational overhead. 
The fixed target set sizes \(s_T{=}3\) and \(s_S{=}5\) were consistently used across all datasets (CIFAR-100, Food-101, DomainNet-Real, and Caltech variants) and yielded stable, reproducible conformal thresholds for both teacher and student predictors.

\section{Identifying regimes where teacher signals no longer help}
\label{app:add-relation}

\begin{figure*}[t]
  \centering
   \begin{subfigure}[t]{0.32\textwidth}
    \includegraphics[width=\linewidth]{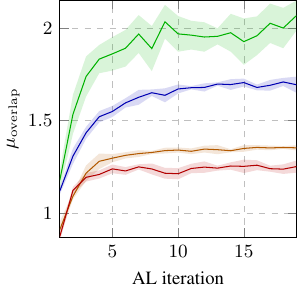}
    \caption{Mean CCMA overlap}
    \label{fig:ccma-overlap}
  \end{subfigure}\hfill
  \begin{subfigure}[t]{0.32\textwidth}
    \includegraphics[width=\linewidth]{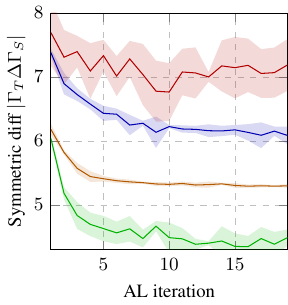}
    \caption{Mean CCMA symmetric difference}
    \label{fig:ccma-symdiff}
  \end{subfigure}\hfill
  \begin{subfigure}[t]{0.32\textwidth}
    \includegraphics[width=\linewidth]{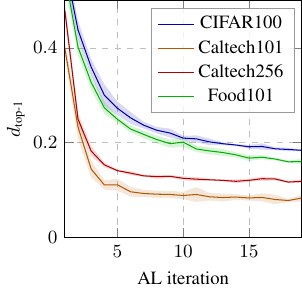}
    \caption{Mean fraction Top-1 Disagreement}
    \label{fig:frac-disagree}
  \end{subfigure}
  \caption{CCMA diagnostic curves across AL iterations. (a) Overlap between teacher and student conformal sets, measuring the agreement in predicted label supports. Higher overlap indicates stronger cross-modal consistency. (b) Symmetric difference between the two sets, capturing the size of their disagreement region. Larger values reflect a persistent teacher–student mismatch. (c) Top-1 disagreement rate, the probability that the teacher and student assign different most-likely classes. These metrics reveal how the informativeness of teacher–student disagreement evolves: datasets like CIFAR100 and Food101 exhibit sustained mismatch (useful for CCMA), whereas Caltech101/256 align quickly, reducing the value of cross-modal uncertainty and shifting the problem toward coverage-based selection.}
  \label{fig:ccma-student-appendix}
\end{figure*}

\begin{figure*}[t]
  \centering
   \begin{subfigure}[t]{0.32\textwidth}
    \includegraphics[width=\linewidth]{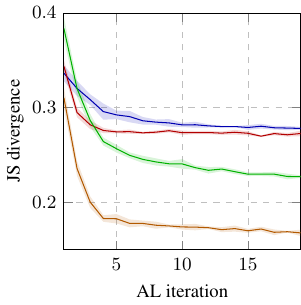}
    \caption{JS Divergence}
    \label{fig:js-full}
  \end{subfigure}\hfill
  \begin{subfigure}[t]{0.32\textwidth}
    \includegraphics[width=\linewidth]{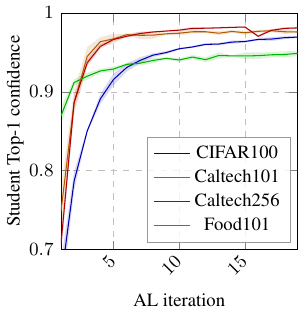}
    \caption{Student Top-1 Confidence}
    \label{fig:student-top1}
  \end{subfigure}\hfill
  \begin{subfigure}[t]{0.32\textwidth}
    \includegraphics[width=\linewidth]{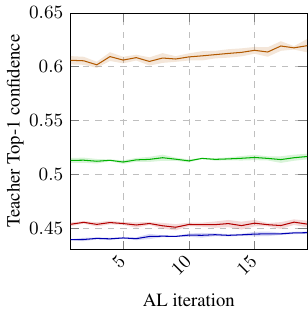}
    \caption{Teacher Top-1 Confidence}
    \label{fig:teacher-top1}
  \end{subfigure}
  \caption{Teacher–student confidence dynamics across AL iterations. (a) JS divergence between teacher and student posteriors restricted to their shared support, quantifying cross-modal disagreement. (b) Student top-1 confidence, showing how quickly the student becomes certain as labeled data accumulates. (c) The teacher's top-1 confidence, which remains relatively stable across iterations since the teacher is frozen. These curves highlight the regimes where CCMA is most effective: datasets like CIFAR100 and Food101 maintain moderate teacher–student disagreement (useful for cross-modal query scoring), whereas Caltech101/256 exhibit rapid confidence saturation and early agreement, reducing the benefit of disagreement-based acquisition.}
  \label{fig:confidence}
\end{figure*}

A core challenge in multimodal active learning is determining when a VLM teacher provides meaningful guidance. 
To answer this, we track four complementary diagnostics across AL iterations:
\textbf{(i)} CCMA set overlap $\mu_{\text{overlap}} = |\Gamma_T \cap \Gamma_S|$ (Fig.~\ref{fig:ccma-overlap}), 
\textbf{(ii)} symmetric difference $|\Gamma_T \triangle \Gamma_S|$ (Fig.~\ref{fig:ccma-symdiff}), 
\textbf{(iii)} Top-1 disagreement $d_{\text{top-1}}$ (Fig.~\ref{fig:frac-disagree}), and 
\textbf{(iv)} JS divergence and predictive confidence trends (Fig.~\ref{fig:confidence}).

\textbf{CIFAR100 and Food101.}  
Across early and mid AL rounds, both disagreement and JS divergence decay gradually 
(Figs.~\ref{fig:frac-disagree} and \ref{fig:js-full}),
indicating that the teacher continues to provide complementary information as the student improves.
Overlap steadily increases while symmetric difference decreases 
(Figs.~\ref{fig:ccma-overlap} and~\ref{fig:ccma-symdiff}),
showing convergence in a \emph{content-aware} manner: the student learns semantics 
that the teacher initially knows better.  
This persistent teacher–student mismatch keeps multimodal uncertainty informative,
enabling CCMA to maintain strong gains throughout the budget.

\textbf{Caltech101 and Caltech256.}  
In contrast, these datasets display a \emph{fast alignment regime}.  
The student rapidly becomes highly confident ($> 0.95$ by $\sim$1k labels; Fig.~\ref{fig:student-top1}),
and the teacher’s confidence remains nearly flat
(Fig.~\ref{fig:teacher-top1}),
suggesting that both models find Caltech classes relatively easy.  
JS divergence and top-1 disagreement collapse early
(Figs.~\ref{fig:js-full} and \ref{fig:frac-disagree}),
meaning the teacher adds almost no new information.  
Here, the symmetric difference stabilizes at a low level, and the overlap saturates quickly  
(Figs.~\ref{fig:ccma-symdiff} and~\ref{fig:ccma-overlap}),
revealing that selection becomes \emph{coverage-dominated} rather than uncertainty-driven.  
Accordingly, clustering-based methods outperform CCMA in later rounds.

These diagnostics expose a previously unaddressed insight:
\emph{VLM guidance is most valuable in regimes with persistent semantic mismatch}.  
Once the teacher is effectively “oracle-like,” relying on teacher–student disagreement harms selection efficiency.
This transforms CCMA into a \textbf{decision rule}:
use cross-modal conformal uncertainty when divergence remains high,  
else transition toward pure coverage as mismatch vanishes.

\section{Impact of hyperparameter choice}
\begin{figure}
  \centering
\includegraphics[width=0.95\linewidth]{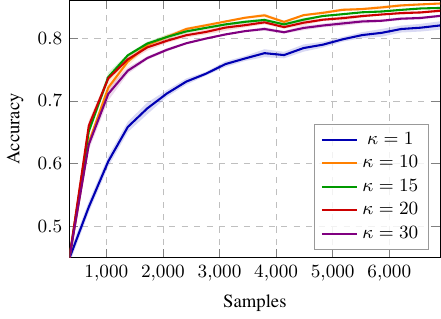}
  \caption{Effect of oversampling factor $\kappa$ on DomainNetReal}
  \label{fig:alpha-factor-domainnetreal}
\end{figure}
Figure~\ref{fig:alpha-factor-domainnetreal} illustrates the sensitivity of CCMA to the oversampling factor $\kappa$ on the DomainNet-Real dataset. Recall that $\kappa$ controls how many top-ranked candidates (by disagreement score) are retained before the final diversity-based clustering. Increasing $\kappa$ enlarges this candidate set, potentially improving selection diversity at the cost of additional computation. 
We observe that performance improves significantly when moving from $\kappa=1$ (no oversampling) to $\kappa=10$, confirming that moderate oversampling effectively stabilizes selection and enhances early learning. However, beyond $\kappa=10$, accuracy gains saturate and slightly decline, indicating diminishing returns and possible noise amplification in large candidate pools. 
This trend contrasts with CIFAR100 and Food101, where higher $\kappa$ values (15–20) yielded more consistent improvements. Overall, $\kappa=10$ provides the best trade-off between accuracy and efficiency for DomainNet-Real, balancing exploration and stability during sample acquisition.

\section{Ablation details}
\label{app:add-ablation}
\begin{table}[t]
\centering
\setlength{\tabcolsep}{3.8pt}
\begin{tabular}{lcccc}
\toprule
Variant & AULC & Final@R & Time/rd (s) & AULC/s \\
\midrule
V1 & 0.859 & 0.915 & 3.015 & 0.285 \\
V2 & 0.828 & 0.899 & 0.780 & 1.061 \\
V3 & 0.835 & 0.906 & 0.512 & 1.629 \\
V4 & 0.809 & 0.896 & 2.929 & 0.276 \\
V5 & 0.793 & 0.885 & 0.583 & 1.360 \\
\bottomrule
\end{tabular}
\caption{Mean over 5 seeds. Final accuracy at the last round (Final@R), AULC computed over rounds, and AULC per second on the CIFAR100 dataset.}
\label{tab:ablation-timing}
\end{table}
Table~\ref{tab:ablation-timing} reports the average performance across five seeds on CIFAR100, comparing several CCMA variants (V1–V5) that differ in their subpooling or diversity-selection configurations. 
We evaluate each variant in terms of the \emph{Area Under the Learning Curve} (AULC), final accuracy at the last round (Final@R), mean computation time per active learning round, and overall efficiency measured as AULC per second.
As shown, V1 achieves the highest overall accuracy and AULC, indicating the strongest acquisition quality, but requires the longest runtime due to full conformal and clustering operations.
Conversely, lightweight variants (e.g., V3 and V5) attain slightly lower accuracy but significantly higher time-normalized efficiency (AULC/s), demonstrating that simplified selection can substantially accelerate active learning without drastic performance loss. 
These results emphasize that CCMA offers a controllable balance between computational cost and selection effectiveness, enabling adaptation to different practical budgets and latency constraints.

\section{All the experiments}

\begin{figure}[!ht]
  \centering
  \begin{subfigure}[t]{\linewidth}
    \includegraphics[width=0.97\linewidth]{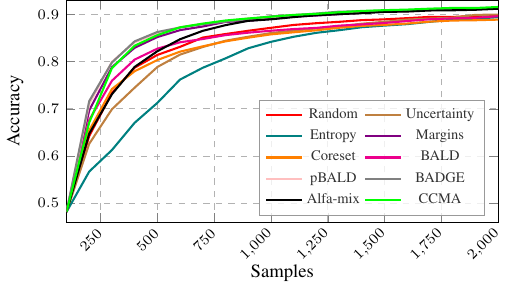}
    \caption{CIFAR100}
    \label{fig:acc-cifar100}
  \end{subfigure}
  \vspace{0.3cm} 
  \begin{subfigure}[t]{\linewidth}
    \includegraphics[width=0.97\linewidth]{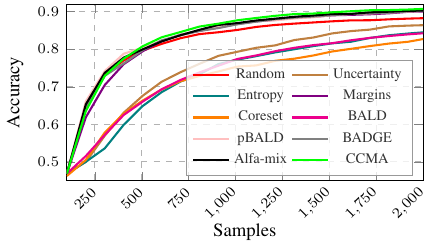}
    \caption{Food101}
    \label{fig:acc-food101}
  \end{subfigure}
  \vspace{0.3cm} 
  \begin{subfigure}[t]{\linewidth}
    \includegraphics[width=0.97\linewidth]{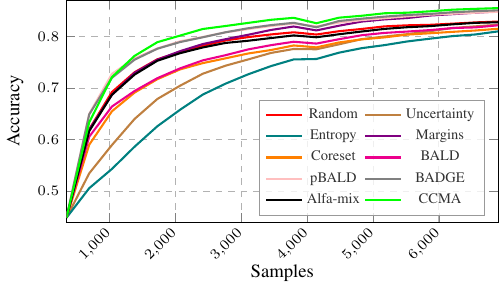}
    \caption{DomainNet-Real}
    \label{fig:acc-domainnetreal}
  \end{subfigure}

  \caption{Test mean accuracy over 5 seeds for CCMA with DINOv2 ViT-g14 as a student model and CLIP ViT-L14 as a teacher model with other AL methods on CIFAR100, Food101, and DomainNet-Real datasets.}
  \label{fig:acc-others-clipv14}
\end{figure}

\begin{figure}[!ht]
  \centering
  \begin{subfigure}[t]{\linewidth}
    \includegraphics[width=0.97\linewidth]{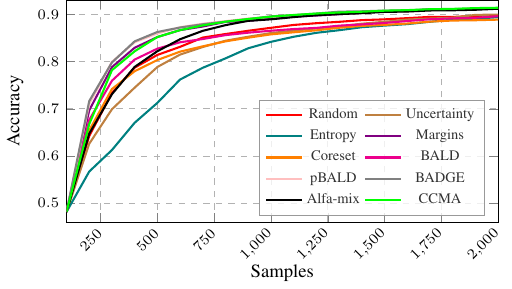}
    \caption{CIFAR100}
    \label{fig:acc-cllipg14-cifar100}
  \end{subfigure}
  \vspace{0.3cm} 
  \begin{subfigure}[t]{\linewidth}
    \includegraphics[width=0.97\linewidth]{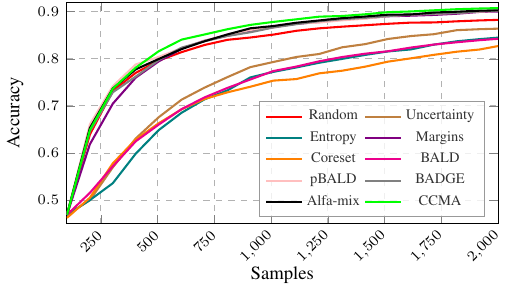}
    \caption{Food101}
    \label{fig:acc-clipg14-food101}
  \end{subfigure}
  \vspace{0.3cm} 
  \begin{subfigure}[t]{\linewidth}
    \includegraphics[width=0.97\linewidth]{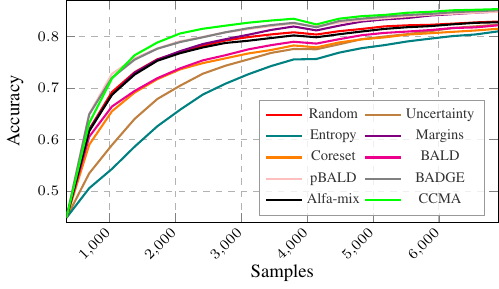}
    \caption{DomainNet-Real}
    \label{fig:acc-clipg14-domainnetreal}
  \end{subfigure}

  \caption{Test mean accuracy over 5 seeds for CCMA with DINOv2 ViT-g14 as a student model and CLIP ViT-g14 as a teacher model with other AL methods on CIFAR100, Food101, and DomainNet-Real datasets.}
  \label{fig:acc-others-clipvitg14}
\end{figure}

\begin{figure}[!ht]
  \centering
  \begin{subfigure}[t]{\linewidth}
    \includegraphics[width=0.97\linewidth]{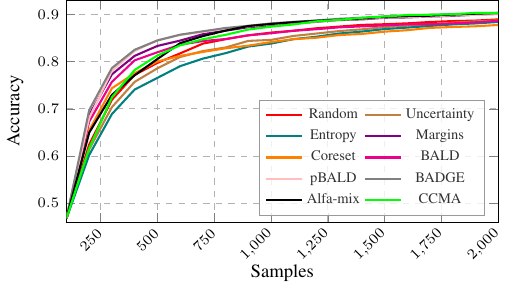}
    \caption{CIFAR100}
    \label{fig:acc-cifar100-dinol14}
  \end{subfigure}
  \vspace{0.3cm} 
  \begin{subfigure}[t]{\linewidth}
    \includegraphics[width=0.97\linewidth]{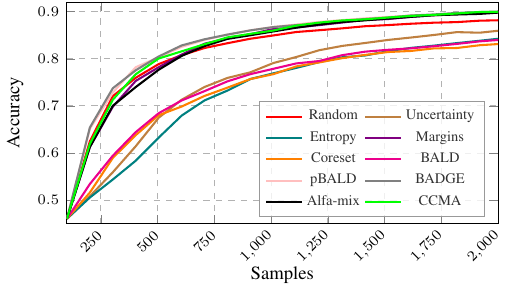}
    \caption{Food101}
    \label{fig:acc-food101-dinol14}
  \end{subfigure}
  \vspace{0.3cm} 
  \begin{subfigure}[t]{\linewidth}
    \includegraphics[width=0.97\linewidth]{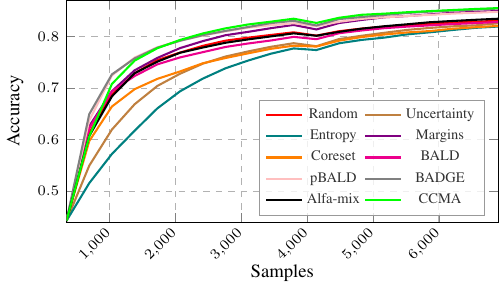}
    \caption{DomainNet-Real}
    \label{fig:acc-domainnetreal-dinol14}
  \end{subfigure}

  \caption{Test mean accuracy over 5 seeds for CCMA with DINOv2 ViT-L14 as a student model and CLIP ViT-L14 as a teacher model with other AL methods on CIFAR100, Food101, and DomainNet-Real datasets.}
  \label{fig:acc-others-clipl14-dinol14}
\end{figure}

\begin{figure}[!ht]
  \centering
  \begin{subfigure}[t]{\linewidth}
    \includegraphics[width=0.97\linewidth]{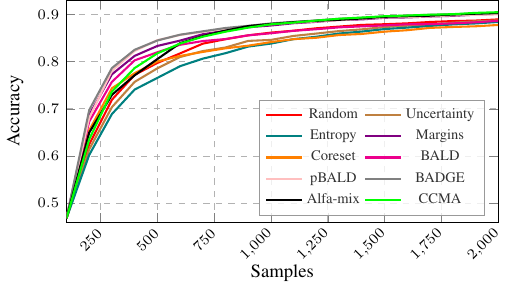}
    \caption{CIFAR100}
    \label{fig:acc-dinol14-cllipg14-cifar100}
  \end{subfigure}
  \vspace{0.3cm} 
  \begin{subfigure}[t]{\linewidth}
    \includegraphics[width=0.97\linewidth]{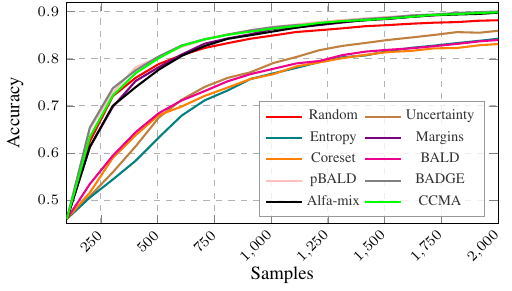}
    \caption{Food101}
    \label{fig:acc-dinol14-clipg14-food101}
  \end{subfigure}
  \vspace{0.3cm} 
  \begin{subfigure}[t]{\linewidth}
    \includegraphics[width=0.97\linewidth]{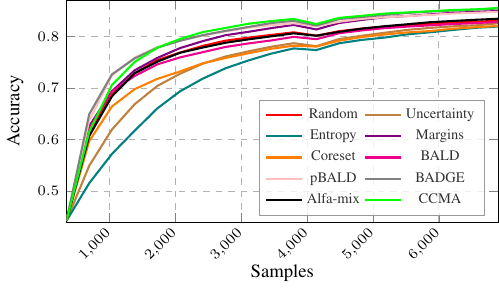}
    \caption{DomainNet-Real}
    \label{fig:acc-dinol14-clipg14-domainnetreal}
  \end{subfigure}

  \caption{Test mean accuracy over 5 seeds for CCMA with DINOv2 ViT-L14 as a student model and CLIP ViT-g14 as a teacher model with other AL methods on CIFAR100, Food101, and DomainNet-Real datasets.}
  \label{fig:acc-others-clipg14-dinol14}
\end{figure}

For completeness, we attempted to include all representative active learning baselines in our experiments. However, two recent clustering-based approaches, \textbf{TypiClust}~\cite{typiclust} and \textbf{ProbCover}~\cite{probcover}, could not be consistently evaluated across all datasets due to inherent constraints in their selection logic. 
In addition, we conducted supplementary experiments by varying the teacher–student backbone configurations, specifically using \textbf{DINOv2 ViT-L/14} and \textbf{DINOv2 ViT-g/14} as students, and \textbf{CLIP ViT-L/14} and \textbf{CLIP ViT-g/14} as teachers, to assess the robustness of CCMA across architectures. The corresponding results are presented in Fig.~\ref{fig:acc-others-clipv14}, \ref{fig:acc-others-clipvitg14}, \ref{fig:acc-others-clipl14-dinol14}, and ~\ref{fig:acc-others-clipg14-dinol14}.

\vspace{0.25em}
\noindent\textbf{TypiClust.}
This method clusters the unlabeled pool in feature space and prioritizes \emph{typical} samples from dense regions for annotation.
While effective at early stages with small budgets, TypiClust progressively depletes its set of available unlabeled “typical” samples, as the clustering density sharply decreases with each iteration.
Consequently, after several rounds (especially on larger datasets such as CIFAR100, Food101, and DomainNet-Real), the algorithm plateaus early and fails to generate new acquisition batches once all cluster centroids have been annotated.
In such cases, TypiClust effectively terminates before reaching the full query budget, preventing a fair comparison in later active learning rounds.

\vspace{0.25em}
\noindent\textbf{ProbCover.}
ProbCover selects samples that maximize coverage of high-uncertainty regions under a probabilistic diversity criterion.
Its selection mechanism relies on computing representative subsets from the entire unlabeled pool, which becomes unstable or infeasible when the pool is too small or lacks sufficient diversity.
This limitation is particularly pronounced in smaller datasets such as Caltech101 and Caltech256, where, after several acquisition rounds, the remaining unlabeled samples no longer satisfy the probabilistic coverage constraints.
As a result, the method halts before completing all iterations, leading to missing curves in our plots for these datasets.

\begin{table*}[!ht]
    \centering
    \caption{Mean accuracy averaged over 5 runs along with the standard deviation at AL iterations $t$ for datasets CIFAR100~\cite{cifar100}, Food101~\cite{food101}, and DomainNet-Real~\cite{domainnetreal} when utilizing the random initialization with DINOv2 ViT-g14 as the feature extractor. \textbf{Bold} values represent the \textbf{first-place} mean accuracy at iteration $t$ with the \underline{second-place} value \underline{underlined}.}
    \vspace{0.5em} 
    \begin{adjustbox}{width=1\textwidth}
    \begin{tabular}{c | c c c c c c c c c c c}
    \toprule 
        $t$ & Random & Uncertainty & Entropy & Margins & BALD & pBALD & Coreset & BADGE & Alfa-mix & ProbCover & \textit{CCMA (ours)} \\
    \midrule 
    \multicolumn{12}{c}{CIFAR100} \\
    \midrule
        1 & $48.0\pm 2.2$ & $48.0\pm 2.2$ & $48.0\pm 2.2$ & $48.0\pm 2.2$ & $48.0\pm 2.2$ & $48.0\pm 2.2$ & $48.0\pm 2.2$ & $48.0\pm 2.2$ & $48.0\pm 2.2$ & $48.0\pm 2.2$ & $48.0\pm 2.2$\\
        2 & $64.2\pm 2.4$ & $62.5\pm 2.4$ & $56.6\pm 3.0$ & $69.6\pm 0.5$ & $67.3\pm 1.0$ & $\underline{71.0}\pm 1.3$ & $64.5\pm 2.1$ & $\mathbf{71.6}\pm 1.1$ & $64.7\pm 2.8$ & $70.4\pm 6.6$ & $67.1\pm 1.5$\\
        3 & $73.3\pm2.1$ & $69.8\pm1.6$ & $61.2\pm3.4$ & $78.7\pm1.0$ & $75.9\pm0.9$ & $\mathbf{79.9}\pm0.7$ & $74.2\pm1.6$ & $\underline{79.7}\pm1.0$ & $73.1\pm2.3$ & $74.4\pm6.6$ & $78.6\pm1.3$\\
        4 & $78.7\pm 1.6$ & $74.4\pm 1.8$ & $67.0\pm 1.2$ & $82.6\pm 1.1$ & $80.4\pm 0.5$ & $\underline{83.9}\pm 0.8$ & $77.9\pm 0.9$ & $\mathbf{84.1}\pm 0.6$ & $78.8\pm 1.4$ & $77.0\pm 5.8$ & $83.4\pm 1.2$\\
        5 & $81.3\pm0.5$ & $78.7\pm1.1$ & $71.2\pm1.1$ & $85.2\pm1.1$ & $82.7\pm0.7$ & $\underline{85.9}\pm0.5$ & $80.3\pm0.9$ & $\mathbf{86.3}\pm0.6$ & $82.2\pm0.5$ & $78.7\pm5.8$ & $85.6\pm0.9$\\
        6 & $83.1\pm0.4$ & $81.5\pm0.9$ & $76.2\pm0.2$ & $86.7\pm0.9$ & $84.1\pm0.6$ & $\mathbf{87.3}\pm0.3$ & $82.1\pm0.9$ & $\underline{87.2}\pm0.3$ & $84.6\pm1.0$ & $79.5\pm5.4$ & $\mathbf{87.3}\pm0.7$\\
        7 & $85.1\pm0.6$ & $83.1\pm0.6$ & $78.6\pm2.3$ & $87.4\pm0.7$ & $84.8\pm0.7$ & $\underline{88.0}\pm0.6$ & $83.2\pm1.2$ & $87.8\pm0.3$ & $86.5\pm0.8$ & $80.3\pm5.2$ & $\mathbf{88.1}\pm0.6$\\
        8 & $85.8\pm 0.7$ & $84.2\pm 0.9$ & $80.7\pm 2.2$ & $87.9\pm 0.7$ & $85.2\pm 0.4$ & $88.4\pm 0.3$ & $84.3\pm 1.0$ & $\underline{88.5}\pm 0.4$ & $87.7\pm 0.8$ & $81.7\pm 4.6$ & $\mathbf{88.7}\pm 0.6$\\
        9 & $86.4\pm0.5$ & $85.2\pm0.6$ & $82.7\pm1.2$ & $88.6\pm0.7$ & $86.1\pm0.3$ & $\underline{89.1}\pm0.2$ & $85.0\pm0.8$ & $89.0\pm0.2$ & $88.6\pm0.4$ & $82.0\pm4.4$ & $\mathbf{89.2}\pm0.5$\\
        10 & $87.0\pm0.4$ & $86.0\pm0.5$ & $84.2\pm1.1$ & $89.1\pm0.6$ & $86.5\pm0.3$ & $89.4\pm0.3$ & $85.7\pm0.6$ & $\underline{89.5}\pm0.3$ & $88.8\pm0.2$ & $83.4\pm3.5$ & $\mathbf{89.6}\pm0.5$\\
        11 & $87.6\pm0.7$ & $86.5\pm0.8$ & $85.2\pm0.7$ & $89.5\pm0.4$ & $86.9\pm0.4$ & $89.7\pm0.1$ & $86.2\pm0.6$ & $\underline{89.8}\pm0.3$ & $89.4\pm0.2$ & $84.0\pm3.6$ & $\mathbf{89.9}\pm0.5$\\
        12 & $88.0\pm0.4$ & $87.0\pm0.9$ & $86.1\pm0.6$ & $89.7\pm0.6$ & $87.1\pm0.6$ & $90.0\pm0.2$ & $86.6\pm0.8$ & $\underline{90.1}\pm0.2$ & $89.8\pm0.2$ & $84.7\pm3.2$ & $\mathbf{90.2}\pm0.3$\\
        13 & $88.4\pm0.4$ & $87.4\pm0.9$ & $86.6\pm0.6$ & $90.0\pm0.3$ & $87.6\pm0.5$ & $\underline{90.1}\pm0.3$ & $87.1\pm1.0$ & $\mathbf{90.5}\pm0.2$ & $90.0\pm0.3$ & $85.2\pm2.9$ & $\mathbf{90.5}\pm0.4$\\
        14 & $88.7\pm0.2$ & $88.0\pm0.6$ & $87.3\pm0.4$ & $90.3\pm0.2$ & $87.9\pm0.5$ & $90.3\pm0.1$ & $87.5\pm0.9$ & $\underline{90.6}\pm0.0$ & $90.2\pm0.2$ & $85.5\pm2.8$ & $\mathbf{90.8}\pm0.4$\\
        15 & $88.9\pm0.3$ & $88.4\pm0.6$ & $87.6\pm0.3$ & $\underline{90.7}\pm0.1$ & $88.2\pm0.2$ & $90.5\pm0.3$ & $87.8\pm0.8$ & $\underline{90.7}\pm0.2$ & $90.5\pm0.2$ & $85.9\pm2.7$ & $\mathbf{90.9}\pm0.4$\\
        16 & $89.2\pm 0.2$ & $88.9\pm 0.8$ & $87.8\pm 0.5$ & $90.7\pm 0.2$ & $88.4\pm 0.3$ & $90.6\pm 0.2$ & $88.3\pm 0.8$ & $\underline{90.8}\pm 0.1$ & $90.6\pm 0.2$ & $86.3\pm 2.6$ & $\mathbf{91.1}\pm 0.4$\\
        17 & $89.5\pm0.1$ & $89.1\pm0.8$ & $88.4\pm0.4$ & $\underline{91.1}\pm0.2$ & $89.0\pm0.2$ & $90.9\pm0.2$ & $88.5\pm0.7$ & $\underline{91.1}\pm0.1$ & $90.8\pm0.2$ & $86.7\pm2.3$ & $\mathbf{91.3}\pm0.3$\\
        18 & $89.4\pm0.1$ & $89.3\pm0.6$ & $88.9\pm0.5$ & $91.0\pm0.2$ & $89.1\pm0.3$ & $91.0\pm0.1$ & $88.7\pm0.7$ & $\underline{91.1}\pm0.1$ & $90.9\pm0.1$ & $86.9\pm2.1$ & $\mathbf{91.4}\pm0.2$\\
        19 & $89.7\pm0.2$ & $89.9\pm0.6$ & $89.1\pm0.5$ & $91.1\pm0.2$ & $89.3\pm0.3$ & $\underline{91.2}\pm0.1$ & $88.8\pm0.8$ & $\underline{91.2}\pm0.1$ & $91.0\pm0.1$ & $87.5\pm1.9$ & $\mathbf{91.4}\pm0.3$\\
        20 & $89.8\pm0.2$ & $89.9\pm0.5$ & $89.4\pm0.1$ & $91.2\pm0.0$ & $89.6\pm0.3$ & $91.2\pm0.2$ & $88.9\pm0.6$ & $\underline{91.3}\pm0.1$ & $91.2\pm0.2$ & $88.0\pm1.8$ & $\mathbf{91.6}\pm0.2$\\
    \midrule
    \multicolumn{12}{c}{Food101} \\
    \midrule 
        1 & $46.8\pm1.9$ & $46.8\pm1.9$ & $46.8\pm1.9$ & $46.8\pm1.9$ & $46.8\pm1.9$ & $46.8\pm1.9$ & $46.8\pm1.9$ & $46.8\pm1.9$ & $46.8\pm1.9$ & $46.8\pm1.9$ & $46.8\pm1.9$ \\
        2 & $63.9\pm2.1$ & $49.9\pm3.2$ & $50.0\pm1.8$ & $61.8\pm2.7$ & $50.8\pm2.3$ & $\underline{66.0}\pm2.3$ & $50.5\pm3.0$ & $64.6\pm1.7$ & $65.3\pm2.1$ & $\mathbf{76.9}\pm0.7$ & $63.5\pm1.0$\\
        3 & $73.1\pm0.3$ & $56.8\pm3.4$ & $53.6\pm2.3$ & $70.4\pm2.1$ & $57.1\pm1.8$ & $\underline{74.1}\pm3.0$ & $57.8\pm2.8$ & $72.8\pm1.4$ & $73.6\pm2.1$ & $\mathbf{80.2}\pm1.1$ & $72.9\pm1.1$\\
        4 & $77.0\pm1.1$ & $63.0\pm3.2$ & $58.6\pm1.5$ & $75.4\pm2.1$ & $62.2\pm2.4$ & $\underline{78.1}\pm2.1$ & $62.8\pm3.7$ & $76.2\pm1.8$ & $77.7\pm1.6$ & $\mathbf{82.0}\pm0.9$ & $\underline{78.1}\pm0.9$\\
        5 & $79.6\pm0.9$ & $67.5\pm2.1$ & $64.7\pm2.6$ & $79.4\pm1.8$ & $66.0\pm2.1$ & $80.2\pm1.9$ & $66.3\pm3.1$ & $79.7\pm1.2$ & $79.8\pm1.4$ & $\mathbf{83.7}\pm0.9$ & $\underline{80.8}\pm1.0$\\
        6 & $81.3\pm1.0$ & $71.3\pm2.1$ & $68.5\pm3.0$ & $81.1\pm1.6$ & $69.3\pm1.7$ & $82.2\pm2.0$ & $69.2\pm3.3$ & $82.3\pm1.4$ & $82.0\pm1.4$ & $\mathbf{84.7}\pm0.8$ & $\underline{83.2}\pm0.6$\\
        7 & $82.8\pm1.0$ & $73.7\pm2.0$ & $71.3\pm2.8$ & $83.7\pm1.4$ & $71.8\pm1.8$ & $83.6\pm0.9$ & $71.4\pm3.4$ & $83.7\pm1.6$ & $83.6\pm1.1$ & $\mathbf{85.7}\pm0.6$ & $\underline{84.6}\pm0.0$\\
        8 & $83.8\pm0.3$ & $76.1\pm1.8$ & $73.0\pm2.4$ & $84.7\pm0.7$ & $73.8\pm2.3$ & $85.0\pm0.8$ & $72.8\pm2.9$ & $84.8\pm0.9$ & $85.1\pm1.1$ & $\underline{86.0}\pm0.5$ & $\mathbf{86.1}\pm0.5$\\
        9 & $84.5\pm0.4$ & $78.2\pm1.9$ & $76.0\pm2.0$ & $85.7\pm0.6$ & $75.6\pm2.1$ & $86.1\pm0.8$ & $74.0\pm2.9$ & $85.6\pm0.9$ & $\underline{86.4}\pm0.4$ & $86.3\pm0.6$ & $\mathbf{86.8}\pm0.6$\\
        10 & $85.1\pm0.1$ & $79.3\pm1.8$ & $77.2\pm2.1$ & $86.6\pm0.9$ & $77.3\pm2.6$ & $86.6\pm0.9$ & $75.3\pm2.3$ & $86.5\pm0.4$ & $\underline{86.8}\pm0.4$ & $86.7\pm0.4$ & $\mathbf{87.7}\pm0.7$\\
        11 & $85.8\pm0.1$ & $80.3\pm1.9$ & $78.1\pm1.4$ & $87.4\pm0.9$ & $78.3\pm2.0$ & $87.5\pm0.5$ & $75.6\pm2.2$ & $87.3\pm0.7$ & $\underline{87.6}\pm0.6$ & $87.3\pm0.3$ & $\mathbf{88.3}\pm0.7$\\
        12 & $86.4\pm0.3$ & $81.0\pm2.1$ & $79.1\pm2.1$ & $87.7\pm0.8$ & $79.4\pm1.6$ & $87.6\pm0.4$ & $76.8\pm2.1$ & $87.8\pm0.7$ & $\underline{88.0}\pm0.5$ & $87.4\pm0.6$ & $\mathbf{88.9}\pm0.3$\\
        13 & $86.8\pm0.2$ & $82.4\pm1.7$ & $79.8\pm2.2$ & $88.3\pm1.0$ & $80.3\pm1.6$ & $88.1\pm0.5$ & $77.4\pm1.7$ & $88.3\pm0.6$ & $\underline{88.6}\pm0.1$ & $87.8\pm0.6$ & $\mathbf{89.3}\pm0.0$\\
        14 & $87.1\pm0.3$ & $83.0\pm1.3$ & $80.7\pm1.4$ & $88.5\pm0.9$ & $81.0\pm1.9$ & $88.4\pm0.3$ & $78.3\pm1.8$ & $88.6\pm0.4$ & $\underline{88.9}\pm0.3$ & $88.3\pm0.4$ & $\mathbf{89.5}\pm0.3$\\
        15 & $87.3\pm0.1$ & $84.1\pm1.4$ & $81.5\pm1.3$ & $89.0\pm0.8$ & $81.5\pm2.1$ & $89.0\pm0.2$ & $79.3\pm1.8$ & $89.0\pm0.4$ & $\underline{89.2}\pm0.1$ & $88.4\pm0.5$ & $\mathbf{89.8}\pm0.2$\\
        16 & $87.4\pm0.4$ & $84.8\pm1.5$ & $81.9\pm1.5$ & $89.2\pm0.8$ & $82.3\pm1.7$ & $\underline{89.3}\pm0.3$ & $80.0\pm1.4$ & $\underline{89.3}\pm0.3$ & $89.1\pm0.2$ & $88.4\pm0.3$ & $\mathbf{90.1}\pm0.2$\\
        17 & $87.7\pm0.2$ & $85.2\pm1.0$ & $82.9\pm1.1$ & $89.2\pm0.8$ & $83.0\pm1.2$ & $89.5\pm0.3$ & $80.8\pm1.4$ & $89.5\pm0.5$ & $\underline{89.8}\pm0.1$ & $88.7\pm0.5$ & $\mathbf{90.3}\pm0.2$\\
        18 & $87.9\pm0.3$ & $86.0\pm0.8$ & $83.7\pm0.8$ & $89.5\pm0.7$ & $83.4\pm1.1$ & $89.7\pm0.3$ & $81.4\pm1.7$ & $89.7\pm0.3$ & $\underline{89.9}\pm0.0$ & $89.0\pm0.7$ & $\mathbf{90.4}\pm0.1$\\
        19 & $88.1\pm0.3$ & $86.2\pm0.8$ & $84.1\pm0.8$ & $89.9\pm0.2$ & $83.9\pm1.4$ & $89.9\pm0.2$ & $81.9\pm1.3$ & $90.0\pm0.2$ & $\underline{90.1}\pm0.2$ & $89.2\pm0.4$ & $\mathbf{90.5}\pm0.1$\\
        20 & $88.2\pm0.3$ & $86.4\pm1.0$ & $84.5\pm0.9$ & $90.0\pm0.1$ & $84.3\pm1.1$ & $90.1\pm0.3$ & $82.8\pm1.1$ & $\underline{90.2}\pm0.3$ & $\underline{90.2}\pm0.1$ & $88.7\pm0.5$ & $\mathbf{90.8}\pm0.2$\\
    \midrule
    \multicolumn{12}{c}{DomainNet-Real} \\
    \midrule 
        1 & $44.7\pm0.8$ & $44.7\pm0.8$ & $44.7\pm0.8$ & $44.7\pm0.8$ & $44.7\pm0.8$ & $44.7\pm0.8$ & $44.7\pm0.8$ & $44.7\pm0.8$ & $44.7\pm0.8$ & $44.7\pm0.8$ & $44.7\pm0.8$\\
        2 & $61.8\pm0.9$ & $53.4\pm1.3$ & $50.4\pm1.6$ & $61.9\pm0.8$ & $60.2\pm0.7$ & $64.9\pm1.2$ & $58.9\pm1.2$ & $64.8\pm1.0$ & $61.6\pm0.5$ & $\mathbf{72.6}\pm0.8$ & $\underline{65.1}\pm0.7$\\
        3 & $69.2\pm0.5$ & $58.9\pm1.8$ & $54.2\pm2.5$ & $68.6\pm0.5$ & $66.4\pm0.6$ & $\underline{72.6}\pm0.8$ & $65.4\pm0.9$ & $71.9\pm0.6$ & $68.6\pm0.6$ & $\mathbf{75.1}\pm0.3$ & $72.1\pm0.8$\\
        4 & $73.0\pm0.7$ & $64.0\pm1.1$ & $58.6\pm2.7$ & $72.9\pm0.5$ & $69.4\pm0.5$ & $75.7\pm0.6$ & $69.1\pm1.1$ & $75.5\pm0.6$ & $72.6\pm0.3$ & $\underline{76.3}\pm0.6$ & $\mathbf{77.7}\pm0.5$\\
        5 & $75.5\pm0.6$ & $67.8\pm1.0$ & $62.5\pm2.4$ & $75.4\pm0.6$ & $71.8\pm0.3$ & $77.6\pm0.5$ & $71.7\pm0.5$ & $77.6\pm0.4$ & $75.2\pm0.2$ & $\underline{77.7}\pm0.6$ & $\mathbf{78.9}\pm0.3$\\
        6 & $76.9\pm0.2$ & $70.4\pm0.7$ & $65.7\pm1.7$ & $77.1\pm0.7$ & $73.8\pm0.2$ & $\underline{79.0}\pm0.3$ & $73.6\pm0.2$ & $78.8\pm0.6$ & $76.7\pm0.5$ & $78.2\pm0.6$ & $\mathbf{80.2}\pm0.4$\\
        7 & $78.1\pm0.3$ & $74.3\pm0.6$ & $68.7\pm1.3$ & $78.4\pm0.5$ & $75.3\pm0.5$ & $\underline{79.9}\pm0.4$ & $74.8\pm0.3$ & $79.8\pm0.2$ & $77.8\pm0.4$ & $78.7\pm0.8$ & $\mathbf{81.5}\pm0.2$\\
        8 & $79.2\pm0.2$ & $74.3\pm0.6$ & $70.8\pm0.8$ & $79.4\pm0.5$ & $76.3\pm0.6$ & $\underline{80.8}\pm0.2$ & $75.7\pm0.4$ & $\underline{80.8}\pm0.1$ & $78.7\pm0.4$ & $78.9\pm0.8$ & $\mathbf{82.0}\pm0.2$\\
        9 & $79.8\pm0.2$ & $75.5\pm0.3$ & $72.6\pm0.7$ & $80.3\pm0.5$ & $77.4\pm0.5$ & $81.4\pm0.2$ & $76.7\pm0.4$ & $\underline{81.6}\pm0.1$ & $79.1\pm0.3$ & $79.4\pm0.7$ & $\mathbf{82.6}\pm0.2$\\
        10 & $80.2\pm0.0$ & $76.8\pm0.3$ & $74.2\pm0.4$ & $81.2\pm0.4$ & $78.3\pm0.7$ & $82.0\pm0.1$ & $77.3\pm0.4$ & $\underline{82.2}\pm0.1$ & $79.6\pm0.2$ & $79.3\pm0.6$ & $\mathbf{83.2}\pm0.2$\\
        11 & $80.8\pm0.1$ & $77.5\pm0.2$ & $75.5\pm0.4$ & $81.8\pm0.2$ & $78.8\pm0.6$ & $82.5\pm0.3$ & $78.2\pm0.4$ & $\underline{82.6}\pm0.2$ & $80.2\pm0.2$ & $79.4\pm0.7$ & $\mathbf{83.6}\pm0.3$\\
        12 & $80.3\pm0.0$ & $77.5\pm0.3$ & $75.6\pm0.3$ & $81.1\pm0.2$ & $78.6\pm0.5$ & $\underline{81.9}\pm0.3$ & $77.8\pm0.5$ & $81.7\pm0.2$ & $79.8\pm0.3$ & $79.3\pm0.5$ & $\mathbf{82.6}\pm0.2$\\
        13 & $81.0\pm0.2$ & $78.4\pm0.4$ & $76.8\pm0.5$ & $82.0\pm0.3$ & $79.4\pm0.4$ & $82.6\pm0.3$ & $78.8\pm0.4$ & $\underline{83.0}\pm0.1$ & $80.4\pm0.1$ & $79.6\pm0.5$ & $\mathbf{83.6}\pm0.1$\\
        14 & $81.4\pm0.2$ & $79.4\pm0.3$ & $77.7\pm0.5$ & $82.8\pm0.2$ & $80.2\pm0.5$ & $83.1\pm0.3$ & $79.4\pm0.5$ & $\underline{83.4}\pm0.1$ & $80.8\pm0.2$ & $79.4\pm0.7$ & $\mathbf{84.0}\pm0.2$\\
        15 & $81.9\pm0.2$ & $79.9\pm0.3$ & $78.3\pm0.4$ & $83.3\pm0.1$ & $80.6\pm0.6$ & $83.5\pm0.3$ & $79.8\pm0.6$ & $\underline{83.8}\pm0.2$ & $81.4\pm0.4$ & $79.4\pm0.5$ & $\mathbf{84.5}\pm0.2$\\
        16 & $82.1\pm0.3$ & $80.4\pm0.4$ & $79.0\pm0.4$ & $83.5\pm0.2$ & $80.9\pm0.4$ & $83.9\pm0.2$ & $80.3\pm0.5$ & $\underline{84.2}\pm0.2$ & $81.7\pm0.5$ & $79.5\pm0.5$ & $\mathbf{84.6}\pm0.2$\\
        17 & $82.2\pm0.1$ & $81.0\pm0.3$ & $79.5\pm0.4$ & $84.0\pm0.0$ & $81.2\pm0.5$ & $84.2\pm0.3$ & $80.6\pm0.5$ & $\underline{84.3}\pm0.1$ & $82.0\pm0.5$ & $79.4\pm0.4$ & $\mathbf{84.9}\pm0.2$\\
        18 & $82.4\pm0.2$ & $81.6\pm0.3$ & $80.0\pm0.4$ & $84.3\pm0.1$ & $81.6\pm0.3$ & $84.4\pm0.2$ & $80.9\pm0.5$ & $\underline{84.6}\pm0.2$ & $82.3\pm0.6$ & $79.6\pm0.6$ & $\mathbf{85.2}\pm0.2$\\
        19 & $82.7\pm0.1$ & $81.9\pm0.2$ & $80.3\pm0.2$ & $84.6\pm0.1$ & $81.7\pm0.3$ & $84.6\pm0.2$ & $81.1\pm0.7$ & $\underline{84.9}\pm0.1$ & $82.6\pm0.6$ & $79.7\pm0.4$ & $\mathbf{85.4}\pm0.1$\\
        20 & $82.8\pm0.1$ & $82.2\pm0.2$ & $80.9\pm0.3$ & $84.8\pm0.0$ & $82.1\pm0.3$ & $84.7\pm0.2$ & $81.4\pm0.6$ & $\underline{85.0}\pm0.1$ & $82.7\pm0.7$ & $79.7\pm0.5$ & $\mathbf{85.5}\pm0.1$\\
    \bottomrule
    \end{tabular}
    \end{adjustbox}
    \label{tab:ccma-accuracy-full}
\end{table*}

\begin{table*}[!ht]
    \centering
    \caption{Test accuracy for our method (\textit{CCMA}) with other AL strategies when utilizing the random initialization with DINOv2 ViT-g14 as the feature extractor. We show AL iterations $t$ for datasets Caltech101 and Caltech256 when utilizing the random initialization. \textbf{Bold} values represent the \textbf{first-place} mean accuracy at iteration $t$ with the \underline{second-place} value \underline{underlined}.}
    \vspace{0.5em} 
    \begin{adjustbox}{width=1\textwidth}
    \begin{tabular}{c | c c c c c c c c c c c}
    \toprule 
        $t$ & Random & Uncertainty & Entropy & Margins & BALD & pBALD & Coreset & BADGE & Alfa-mix & Typiclust & \textit{CCMA (ours)} \\
    \midrule
    \multicolumn{12}{c}{Caltech-256} \\
    \midrule
        1 & $57.4\pm1.5$ & $57.4\pm1.5$ & $57.4\pm1.5$ & $57.4\pm1.5$ & $57.4\pm1.5$ & $57.4\pm1.5$ & $57.4\pm1.5$ & $57.4\pm1.5$ & $57.4\pm1.5$ & $57.4\pm1.5$ & $57.4\pm1.5$\\
        2 & $76.4\pm1.4$ & $70.0\pm1.6$ & $66.3\pm1.6$ & $78.9\pm1.0$ & $78.1\pm0.6$ & $\underline{82.9}\pm1.1$ & $81.5\pm1.1$ & $\mathbf{83.9}\pm0.8$ & $75.8\pm1.0$ & $75.3\pm1.8$ & $82.2\pm1.0$\\
        3 & $83.8\pm1.4$ & $77.7\pm2.8$ & $72.0\pm1.5$ & $86.0\pm0.9$ & $85.1\pm0.5$ & $\underline{90.0}\pm0.3$ & $87.0\pm0.5$ & $89.7\pm0.6$ & $83.7\pm0.7$ & $84.4\pm1.0$ & $\mathbf{90.8}\pm0.8$\\
        4 & $87.9\pm1.0$ & $83.8\pm1.3$ & $76.5\pm1.9$ & $89.6\pm0.5$ & $87.8\pm0.7$ & $\underline{92.3}\pm0.5$ & $89.3\pm0.7$ & $92.0\pm0.3$ & $88.5\pm0.9$ & $88.4\pm1.3$ & $\mathbf{93.1}\pm0.4$\\
        5 & $90.4\pm0.7$ & $87.2\pm1.3$ & $80.8\pm1.9$ & $92.1\pm0.5$ & $90.1\pm0.7$ & $93.5\pm0.2$ & $90.7\pm0.7$ & $\underline{93.8}\pm0.4$ & $90.6\pm0.7$ & $91.3\pm1.5$ & $\mathbf{94.1}\pm0.2$\\
        6 & $92.0\pm0.3$ & $89.9\pm1.4$ & $84.4\pm1.4$ & $93.8\pm0.3$ & $91.8\pm0.8$ & $94.5\pm0.2$ & $92.0\pm0.6$ & $\underline{94.7}\pm0.5$ & $92.0\pm0.4$ & $92.3\pm0.5$ & $\mathbf{94.8}\pm0.1$\\
        7 & $92.7\pm0.4$ & $92.6\pm1.0$ & $87.5\pm1.2$ & $94.8\pm0.4$ & $93.1\pm0.7$ & $\underline{95.2}\pm0.3$ & $92.7\pm0.7$ & $\mathbf{95.6}\pm0.4$ & $92.9\pm0.4$ & $93.1\pm0.2$ & $\underline{95.2}\pm0.1$\\
        8 & $93.4\pm0.3$ & $93.9\pm0.8$ & $89.9\pm1.3$ & $95.5\pm0.1$ & $94.1\pm0.4$ & $\underline{95.8}\pm0.2$ & $93.2\pm0.8$ & $\mathbf{96.2}\pm0.3$ & $93.8\pm0.5$ & $93.5\pm0.3$ & $95.4\pm0.1$\\
        9 & $94.0\pm0.3$ & $95.1\pm0.5$ & $91.7\pm1.0$ & $\underline{96.3}\pm0.1$ & $94.8\pm0.5$ & $96.2\pm0.2$ & $93.9\pm0.6$ & $\mathbf{97.0}\pm0.3$ & $94.6\pm0.7$ & $94.1\pm0.3$ & $95.8\pm0.0$\\
        10 & $94.4\pm0.2$ & $96.1\pm0.3$ & $93.4\pm0.7$ & $\underline{96.9}\pm0.2$ & $95.4\pm0.2$ & $96.6\pm0.2$ & $94.4\pm0.5$ & $\mathbf{97.3}\pm0.3$ & $95.5\pm0.9$ & $94.3\pm0.3$ & $95.9\pm0.0$\\
        11 & $94.6\pm0.1$ & $96.7\pm0.2$ & $94.3\pm0.8$ & $\underline{97.1}\pm0.1$ & $95.9\pm0.3$ & $97.0\pm0.3$ & $94.9\pm0.3$ & $\mathbf{97.7}\pm0.3$ & $95.7\pm0.9$ & $94.5\pm0.3$ & $96.1\pm0.0$\\
        12 & $94.8\pm0.1$ & $97.2\pm0.3$ & $95.5\pm0.7$ & $\underline{97.5}\pm0.2$ & $96.5\pm0.2$ & $97.3\pm0.1$ & $95.1\pm0.3$ & $\mathbf{98.0}\pm0.2$ & $96.0\pm0.9$ & $94.7\pm0.2$ & $96.3\pm0.1$\\
        13 & $95.0\pm0.2$ & $97.6\pm0.3$ & $96.5\pm0.6$ & $\underline{97.9}\pm0.2$ & $97.0\pm0.2$ & $97.6\pm0.1$ & $95.6\pm0.2$ & $\mathbf{98.2}\pm0.1$ & $96.5\pm1.1$ & $94.9\pm0.2$ & $96.4\pm0.0$\\
        14 & $95.1\pm0.2$ & $\underline{98.0}\pm0.1$ & $97.0\pm0.6$ & $\underline{98.0}\pm0.2$ & $97.3\pm0.2$ & $\underline{98.0}\pm0.1$ & $95.7\pm0.3$ & $\mathbf{98.4}\pm0.1$ & $96.7\pm1.0$ & $95.0\pm0.1$ & $96.6\pm0.0$\\
        15 & $95.3\pm0.1$ & $\underline{98.0}\pm0.2$ & $97.1\pm0.4$ & $\mathbf{98.1}\pm0.1$ & $97.5\pm0.3$ & $97.9\pm0.1$ & $96.0\pm0.2$ & $97.8\pm0.2$ & $97.0\pm0.8$ & $95.1\pm0.0$ & $96.7\pm0.0$\\
        16 & $94.8\pm0.1$ & $97.4\pm0.2$ & $97.2\pm0.3$ & $\underline{97.7}\pm0.1$ & $97.4\pm0.3$ & $97.6\pm0.1$ & $95.5\pm0.2$ & $\mathbf{97.9}\pm0.2$ & $96.6\pm0.5$ & $94.8\pm0.9$ & $96.3\pm0.1$\\
        17 & $95.3\pm0.1$ & $\underline{98.5}\pm0.2$ & $98.0\pm0.3$ & $\mathbf{98.8}\pm0.1$ & $98.0\pm0.1$ & $98.3\pm0.0$ & $96.2\pm0.1$ & $\mathbf{98.8}\pm0.1$ & $97.4\pm0.5$ & $95.4\pm0.1$ & $96.7\pm0.0$\\
        18 & $95.4\pm0.2$ & $98.7\pm0.2$ & $98.4\pm0.3$ & $\underline{99.0}\pm0.1$ & $98.4\pm0.2$ & $98.6\pm0.0$ & $96.3\pm0.2$ & $\mathbf{99.1}\pm0.2$ & $97.8\pm0.5$ & $95.5\pm0.1$ & $96.9\pm0.1$\\
        19 & $95.6\pm0.1$ & $\underline{98.8}\pm0.2$ & $98.7\pm0.3$ & $\mathbf{99.2}\pm0.1$ & $98.7\pm0.1$ & $98.7\pm0.0$ & $96.5\pm0.3$ & $\mathbf{99.2}\pm0.1$ & $98.0\pm0.4$ & $95.6\pm0.0$ & $97.1\pm0.0$\\
        20 & $95.8\pm0.1$ & $\underline{99.1}\pm0.1$ & $98.9\pm0.3$ & $\mathbf{99.3}\pm0.0$ & $98.9\pm0.0$ & $98.8\pm0.1$ & $96.8\pm0.1$ & $\mathbf{99.3}\pm0.0$ & $98.3\pm0.4$ & $95.7\pm0.1$ & $97.2\pm0.1$\\
    \midrule
    \multicolumn{12}{c}{Caltech-101} \\
    \midrule
        1 & $61.5\pm2.0$ & $61.5\pm2.0$ & $61.5\pm2.0$ & $61.5\pm2.0$ & $61.5\pm2.0$ & $61.5\pm2.0$ & $61.5\pm2.0$ & $61.5\pm2.0$ & $61.5\pm2.0$ & $61.5\pm2.0$ & $61.5\pm2.0$\\
        2 & $77.0\pm1.4$ & $75.1\pm3.1$ & $71.0\pm2.2$ & $82.3\pm2.5$ & $82.2\pm2.0$ & $85.2\pm1.0$ & $\mathbf{88.6}\pm1.4$ & $\underline{86.7}\pm1.1$ & $76.9\pm2.7$ & $76.4\pm2.0$ & $75.7\pm1.7$\\
        3 & $84.9\pm2.2$ & $83.5\pm1.2$ & $79.7\pm1.5$ & $88.5\pm0.8$ & $88.5\pm1.3$ & $\mathbf{90.6}\pm0.5$ & $\underline{90.1}\pm0.5$ & $\mathbf{90.6}\pm0.8$ & $84.1\pm1.8$ & $83.7\pm1.2$ & $82.2\pm0.8$\\
        4 & $89.4\pm0.9$ & $87.8\pm0.6$ & $85.8\pm1.5$ & $90.4\pm0.8$ & $90.3\pm0.6$ & $\underline{92.1}\pm0.7$ & $90.8\pm0.7$ & $\mathbf{92.2}\pm0.7$ & $88.0\pm1.5$ & $88.3\pm0.4$ & $86.1\pm0.6$\\
        5 & $90.9\pm0.3$ & $89.6\pm0.5$ & $88.3\pm1.4$ & $92.5\pm0.8$ & $91.0\pm0.5$ & $\underline{93.6}\pm0.7$ & $90.9\pm0.7$ & $\mathbf{94.1}\pm0.8$ & $89.8\pm1.5$ & $90.4\pm1.1$ & $90.0\pm0.7$\\
        6 & $92.5\pm0.5$ & $91.5\pm0.6$ & $89.8\pm0.8$ & $94.1\pm0.7$ & $91.9\pm0.7$ & $\underline{94.7}\pm0.7$ & $91.4\pm0.6$ & $\mathbf{95.2}\pm0.5$ & $91.1\pm0.7$ & $91.7\pm0.6$ & $91.8\pm0.9$\\
        7 & $93.6\pm0.7$ & $93.2\pm0.4$ & $91.2\pm0.6$ & $94.9\pm0.6$ & $93.6\pm0.6$ & $\underline{95.5}\pm0.7$ & $91.5\pm0.6$ & $\mathbf{95.8}\pm0.7$ & $92.5\pm0.5$ & $92.6\pm0.9$ & $92.8\pm0.9$\\
        8 & $94.6\pm0.7$ & $94.5\pm0.5$ & $92.6\pm1.1$ & $95.5\pm0.9$ & $94.7\pm0.4$ & $\underline{96.0}\pm0.7$ & $91.6\pm0.8$ & $\mathbf{96.3}\pm0.8$ & $92.7\pm0.6$ & $93.4\pm0.4$ & $94.8\pm0.1$\\
        9 & $95.0\pm0.3$ & $95.5\pm0.5$ & $94.2\pm1.1$ & $95.9\pm1.1$ & $95.9\pm0.5$ & $\underline{96.5}\pm0.7$ & $92.2\pm0.8$ & $\mathbf{96.6}\pm0.6$ & $93.3\pm0.8$ & $93.9\pm0.3$ & $93.8\pm0.5$\\
        10 & $95.3\pm0.3$ & $96.1\pm0.5$ & $95.1\pm0.7$ & $96.3\pm1.0$ & $96.7\pm0.7$ & $\underline{96.8}\pm0.8$ & $92.7\pm0.8$ & $\mathbf{96.9}\pm0.7$ & $94.0\pm1.2$ & $94.4\pm0.6$ & $94.5\pm0.5$\\
        11 & $95.5\pm0.3$ & $96.3\pm0.7$ & $95.8\pm0.8$ & $96.6\pm1.0$ & $\underline{97.0}\pm0.6$ & $\underline{97.0}\pm0.5$ & $92.7\pm0.9$ & $\mathbf{97.2}\pm0.7$ & $95.0\pm1.7$ & $94.8\pm0.5$ & $95.1\pm0.4$\\
        12 & $95.6\pm0.4$ & $96.7\pm0.7$ & $96.4\pm0.8$ & $\underline{97.0}\pm0.9$ & $\mathbf{97.4}\pm0.6$ & $\mathbf{97.4}\pm0.6$ & $93.1\pm1.0$ & $\mathbf{97.4}\pm0.6$ & $95.9\pm1.5$ & $94.9\pm0.5$ & $95.2\pm0.6$\\
        13 & $95.8\pm0.4$ & $96.9\pm0.6$ & $96.8\pm0.6$ & $97.2\pm0.9$ & $\mathbf{97.5}\pm0.6$ & $\mathbf{97.5}\pm0.6$ & $94.0\pm0.9$ & $\underline{97.4}\pm0.8$ & $96.6\pm1.2$ & $95.3\pm0.4$ & $95.6\pm0.3$\\
        14 & $96.0\pm0.3$ & $97.0\pm0.2$ & $97.1\pm0.4$ & $97.4\pm0.8$ & $\mathbf{97.9}\pm0.4$ & $\underline{97.6}\pm0.4$ & $94.1\pm0.6$ & $97.5\pm0.7$ & $97.4\pm0.4$ & $95.7\pm0.2$ & $95.7\pm0.3$\\
        15 & $96.1\pm0.4$ & $97.3\pm0.5$ & $97.1\pm0.3$ & $97.5\pm0.7$ & $\mathbf{98.0}\pm0.4$ & $\mathbf{98.0}\pm0.6$ & $94.3\pm0.6$ & $\underline{97.8}\pm0.8$ & $\mathbf{98.0}\pm0.3$ & $95.7\pm0.1$ & $95.9\pm0.2$\\
        16 & $96.2\pm0.2$ & $97.2\pm0.5$ & $97.2\pm0.3$ & $97.7\pm0.6$ & $\mathbf{98.1}\pm0.3$ & $\mathbf{98.1}\pm0.6$ & $94.4\pm0.5$ & $\underline{97.8}\pm0.8$ & $\mathbf{98.1}\pm0.3$ & $95.8\pm0.1$ & $96.6\pm0.2$\\
        17 & $96.3\pm0.4$ & $97.6\pm0.5$ & $97.3\pm0.3$ & $97.9\pm0.5$ & $\mathbf{98.4}\pm0.3$ & $\underline{98.3}\pm0.6$ & $94.7\pm0.7$ & $98.0\pm0.8$ & $\mathbf{98.4}\pm0.1$ & $96.1\pm0.1$ & $96.4\pm0.2$\\
        18 & $96.6\pm0.2$ & $98.0\pm0.8$ & $97.5\pm0.2$ & $98.2\pm0.5$ & $\underline{98.4}\pm0.5$ & $\underline{98.4}\pm0.7$ & $94.9\pm0.6$ & $98.2\pm0.7$ & $\mathbf{98.5}\pm0.0$ & $96.3\pm0.2$ & $96.5\pm0.2$\\
        19 & $96.7\pm0.1$ & $98.2\pm0.6$ & $97.6\pm0.2$ & $98.4\pm0.3$ & $\underline{98.5}\pm0.5$ & $\underline{98.5}\pm0.8$ & $95.4\pm0.7$ & $\underline{98.5}\pm0.3$ & $\mathbf{98.6}\pm0.1$ & $96.5\pm0.1$ & $96.5\pm0.1$\\
        20 & $96.8\pm0.3$ & $98.3\pm0.6$ & $98.0\pm0.4$ & $98.4\pm0.5$ & $\underline{98.5}\pm0.5$ & $\underline{98.5}\pm0.7$ & $95.6\pm0.5$ & $\underline{98.5}\pm0.4$ & $\mathbf{98.6}\pm0.0$ & $96.6\pm0.1$ & $96.7\pm0.2$\\
    \bottomrule
    \end{tabular}
    \end{adjustbox}
    \label{tab:ccma-caltech}
\end{table*}
\end{document}